\documentclass[10pt,twocolumn,letterpaper]{article}

\usepackage[pagenumbers]{cvpr_2025}              % To produce the CAMERA-READY version
\definecolor{cvprblue}{rgb}{0.21,0.49,0.74}
\usepackage[pagebackref,breaklinks,colorlinks,allcolors=cvprblue]{hyperref}
\usepackage{multirow}
\usepackage{makecell}
\usepackage{algorithm}
\usepackage{algorithmic}
\usepackage{bbding}
\def\paperID{4726} % *** Enter the Paper ID here
\def\confName{CVPR}
\def\confYear{2025}

\title{DriveDreamer4D: World Models Are Effective Data Machines \\ for 4D Driving Scene Representation}

\author{Guosheng Zhao\footnotemark[1]~\textsuperscript{\rm \ 1, 2}~~~~~~Chaojun Ni\footnotemark[1]~\textsuperscript{\rm \ 1, 4}~~~~~~Xiaofeng Wang\footnotemark[1]~\textsuperscript{\rm \ 1, 2}~~~~~~Zheng Zhu\footnotemark[1]~\textsuperscript{\rm \ 1}\textsuperscript{\Envelope}
\\
~~~~~~Xueyang Zhang\textsuperscript{\rm 3}~~~~~~
~~~~~~Yida Wang\textsuperscript{\rm 3}~~~~~~ 
~~~~~~Guan Huang\textsuperscript{\rm 1}~~~~~~ 
Xinze Chen\textsuperscript{\rm 1}
~~~~Boyuan Wang\textsuperscript{\rm 1, 2}
\\
~~~~Youyi Zhang\textsuperscript{\rm 5}
~~~~Wenjun Mei\textsuperscript{\rm 4}
~~~~Xingang Wang\textsuperscript{\rm 2}\textsuperscript{\Envelope}\\
\textsuperscript{\rm 1}GigaAI
~ ~ \textsuperscript{\rm 2}Institute of Automation, Chinese Academy of Sciences
\\
~ ~ \textsuperscript{\rm 3}
Li Auto Inc.
~ ~ \textsuperscript{\rm 4}Peking University
~ ~ \textsuperscript{\rm 5} 
Technical University o
f Munich
\\
\small{Project Page: \url{https://drivedreamer4d.github.io}}
}

\begin{document}
% \maketitle
\twocolumn[{%
% \vspace{-1em}
\maketitle
\vspace{-3em}
\begin{center}
% \vspace{-2em}
\centering
\resizebox{1\linewidth}{!}{
\includegraphics{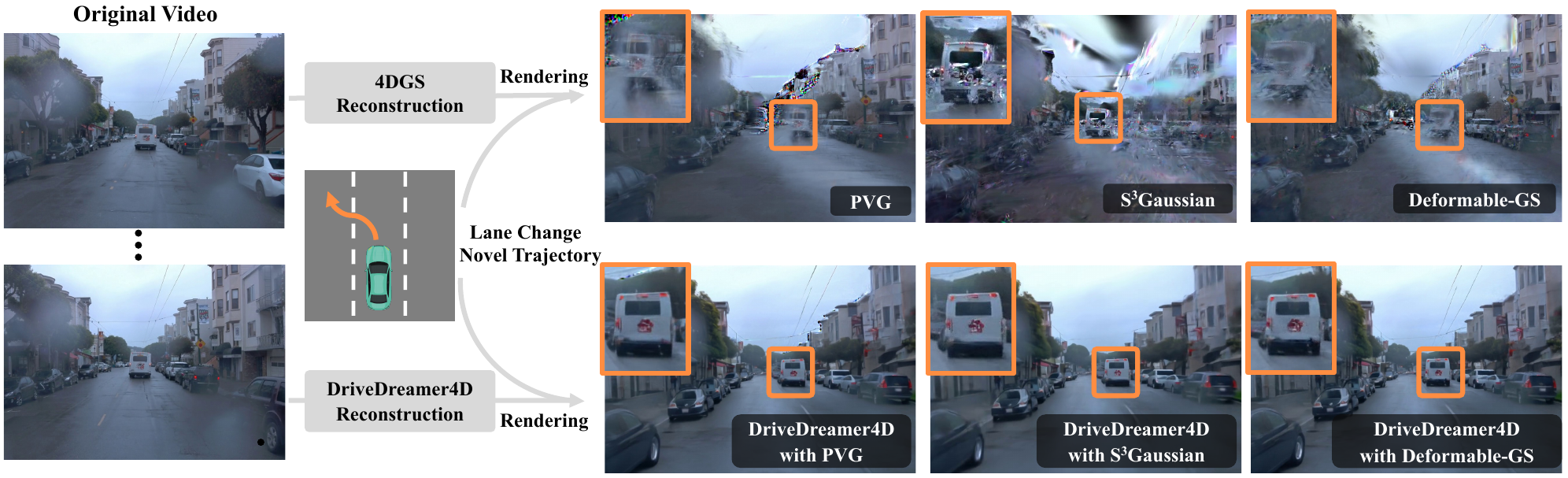
}}
% \vspace{-0.8em}
\captionof{figure}{Previous 4D Gaussian Splatting methods (e.g., PVG \cite{pvg}, $\text{S}^3\text{Gaussian}$ \cite{s3gaussian}, Deformable-GS \cite{deformablegs}) face challenges in rendering novel trajectories, such as lane change. \textit{DriveDreamer4D} addresses this by enhancing 4D driving scene representation via integrating priors from world models, significantly improving rendering quality under complex scenarios and novel trajectory viewpoints.}
\label{fig:main}
\end{center}}]

\renewcommand{\thefootnote}{\fnsymbol{footnote}}
\footnotetext[1]{These authors contributed equally to this work.
\textsuperscript{\Envelope}Corresponding authors: Zheng Zhu, zhengzhu@ieee.org, Xingang Wang, xingang.wang@ia.ac.cn.} 

\begin{abstract}
Closed-loop simulation is essential for advancing end-to-end autonomous driving systems. Contemporary sensor simulation methods, such as NeRF and 3DGS, rely predominantly on conditions closely aligned with training data distributions, which are largely confined to forward-driving scenarios. Consequently, these methods face limitations when rendering complex maneuvers (e.g., lane change, acceleration, deceleration). Recent advancements in autonomous-driving world models have demonstrated the potential to generate diverse driving videos. However, these approaches remain constrained to 2D video generation, inherently lacking the spatiotemporal coherence required to capture intricacies of dynamic driving environments.
In this paper, we introduce \textit{DriveDreamer4D}, which enhances 4D driving scene representation leveraging world model priors. 
Specifically, we utilize the world model as a data machine to synthesize novel trajectory videos, where structured conditions are explicitly leveraged to control the spatial-temporal consistency of traffic elements. Besides, the cousin data training strategy is proposed to facilitate merging real and synthetic data for optimizing 4DGS. To our knowledge, \textit{DriveDreamer4D} is the first to utilize video generation models for improving 4D reconstruction in driving scenarios.
Experimental results reveal that \textit{DriveDreamer4D} significantly enhances generation quality under novel trajectory views, achieving a relative improvement in FID  by 32.1\%, 46.4\%, and 16.3\% compared to PVG, $\text{S}^3$Gaussian, and Deformable-GS. Moreover, \textit{DriveDreamer4D} markedly enhances the spatiotemporal coherence of driving agents, which is verified by a comprehensive user study and the relative increases of 22.6\%, 43.5\%, and 15.6\% in the NTA-IoU metric.

\end{abstract}

\section{Introduction}
End-to-end planning \cite{stp3,uniad,vad}, which directly maps sensor inputs to control signals, is among the most critical and promising tasks in autonomous driving. However, current open-loop evaluations are inadequate for accurately assessing end-to-end planning algorithms, highlighting an urgent need for enhanced evaluation methods \cite{li2024ego,admlp,xld}. A compelling solution 
lies in closed-loop evaluations within real-world scenarios, which require retrieving sensor data from arbitrarily specified viewpoints. This necessitates constructing a 4D driving scene representation capable of reconstructing complex, dynamic driving environments.

Closed-loop simulation in driving environments predominantly relies on scene reconstruction techniques such as Neural Radiance Fields (NeRF) \cite{nerf,emernerf,unisim,streetsurf} and 3D Gaussian Splatting (3DGS) \cite{3dgs,streetgaussian,s3gaussian,omnire}, which are inherently limited by the density of input data. Specifically, these methods can render scenes effectively only under conditions closely aligned with their training data distributions—primarily forward-driving scenarios—and struggle to perform accurately during complex maneuvers (see Fig.~\ref{fig:main}). To mitigate these limitations, methods like SGD \cite{sgd} and GGS \cite{han2024ggs} leverage generative models to extend the range of training viewpoints. However, these approaches primarily supplement sparse image data or static background elements, falling short of modeling the intricacies of dynamic, interactive driving scenes. Recently, advancements in autonomous driving world models \cite{drivedreamer,drivedreamer2,worlddreamer,gaia,drivewm,vista} have introduced the capability to generate diverse, command-aligned video viewpoints, offering renewed promise for closed-loop simulation in autonomous driving. Nonetheless, these models remain constrained to 2D videos, lacking the spatial-temporal coherence essential for accurately modeling complex driving scenarios.

In this paper, we introduce \textit{DriveDreamer4D}, which improves 4D driving scene representation by integrating priors from autonomous driving world models. Our approach utilizes an autonomous driving world model \cite{drivedreamer2} as a generative engine, synthesizing novel trajectory video data that densifies real-world driving datasets for enhanced training. Notably, we propose the Novel Trajectory Generation Module (NTGM) to generate diverse structured traffic conditions, and \textit{DriveDreamer4D} applies these conditions to independently regulate the motion dynamics of foreground and background elements in complex driving environments. These conditions undergo view projection synchronized with vehicle maneuvers, ensuring that the synthesized data adheres to the spatiotemporal constraints. 
Subsequently, the Cousin Data Training Strategy (CDTS) is proposed to merge temporal-aligned real and synthetic data for training 4DGS. In CDTS, a regularization loss is further incorporated to ensure perceptual coherence. To the best of our knowledge, \textit{DriveDreamer4D} is the first framework to harness video generation models for elevating 4D scene reconstruction quality in autonomous driving, providing richly varied viewpoint data for scenarios including lane change, acceleration, and deceleration.  As shown in Fig.~\ref{fig:main}, experiment results demonstrate that \textit{DriveDreamer4D} significantly enhances generation fidelity for novel trajectory viewpoints, achieving a relative improvement in FID  by 32.1\%, 46.4\%, and 16.3\% compared to PVG \cite{pvg}, $\text{S}^3$Gaussian \cite{s3gaussian}, and Deformable-GS \cite{deformablegs}. Besides, \textit{DriveDreamer4D} fortifies the spatiotemporal coherence between foreground and background elements, with respective increases of 22.6\%, 43.5\%, and 15.6\% in the NTA-IoU metric. Furthermore, a comprehensive user study confirms that the average win rate of \textit{DriveDreamer4D} exceeds 80\%, compared to three baselines.

The primary contributions of this work are as follows: (1) We present \textit{DriveDreamer4D}, the first framework to leverage world model priors for advancing 4D scene reconstruction in autonomous driving. (2) The NTGM is proposed to automate the generation of structured conditions, allowing \textit{DriveDreamer4D} to create novel trajectory videos with complex maneuvers while ensuring spatial-temporal consistency. Additionally, the CDTS is introduced to merge temporal-aligned real and synthetic data for training 4DGS, using a regularization loss to maintain perceptual coherence.
(3) We perform comprehensive experiments to validate that \textit{DriveDreamer4D} notably enhances generation quality across novel trajectory viewpoints, as well as the spatiotemporal coherence of driving scene elements.

\section{Related Work}

\subsection{Driving Scene Representation}
NeRF and 3DGS have emerged as leading approaches for 3D scene representation. NeRF models \cite{nerf,mipnerf,zipnerf,ngp} continuous volumetric scenes using multi-layer perceptron (MLP) networks, enabling highly detailed scene reconstructions with remarkable rendering quality. More recently, 3DGS \cite{3dgs,mipgs} introduces an innovative method by defining a set of anisotropic Gaussians in 3D space, leveraging adaptive density control to achieve high-quality renderings from sparse point cloud inputs.
Several works have extended NeRF \cite{unisim,emernerf,neo360,lu2023urban,urbannerf,blocknerf,streetsurf,ucnerf} or 3DGS \cite{streetgaussian,drivinggaussian,pvg,sgd,s3gaussian,omnire,hogaussian} to autonomous driving scenarios. Given the dynamic nature of driving environments, there has also been significant effort in modeling 4D driving scene representations. Some approaches encode time as an additional input to parameterize 4D scenes \cite{attal2023hyperreel,kplane,li2021neural,lin2022efficient,hypernerf,nerfplayer,s3gaussian}, while others represent scenes as a composition of moving object models alongside a static background model \cite{unisim,panopticnerf,ost2021neural,Neurad,mars,snerf}. Despite these advancements, methods based on NeRF and 3DGS face limitations tied to the density of input data. These techniques can only render scenes effectively when sensor data closely matches the training data distribution, which is typically confined to forward-driving scenarios.

\subsection{World Models}
The world model module predicts possible future world states as a function of imagined action sequences proposed by the actor \cite{lecun2022jepa,zhu2024sora}. Approaches such as \cite{worlddreamer,yan2021videogpt,pixeldance,emuvideo,gupta2023photorealistic,svd,ho2022imagen,ma2024latte,ho2022video,videoldm,kondratyuk2023videopoet,yang2024cogvideox,hong2022cogvideo,xiang2024pandora,egovid} simulate environments through video generation controlled by free-text actions. At the forefront of this evolution is Sora \cite{videoworldsimulators2024}, which leverages advanced generative techniques to produce intricate visual sequences that respect the fundamental laws of physics. This ability to deeply understand and simulate the environment not only improves video generation quality but also has substantial implications for real-world driving scenarios. Autonomous driving world models \cite{drivedreamer,drivedreamer2,drivewm,gaia,vista,yang2024drivearena} employ predictive methodologies to interpret driving environments, thereby generating realistic driving scenarios and learning key driving elements and policies from video data. Although these models successfully produce diverse driving video data conditioned on complex driving actions, they remain limited to 2D outputs and lack the spatial-temporal coherence needed to accurately capture the complexities of dynamic driving environments. 

\subsection{Diffusion Prior for 3D Representation}
Constructing comprehensive 3D scenes from limited observations demands generative prior, particularly for unseen areas. Earlier studies distill the knowledge from text-to-image diffusion models \cite{sd,sdxl,saharia2022photorealistic,ramesh2022hierarchical} into a 3D representation model. Specifically, the Score Distillation Sampling (SDS) \cite{dreamfusion,magic3d,reconfusion} is adopted to synthesize a 3D object from the text prompt. Furthermore, to enhance 3D consistency, several approaches extend the multi-view diffusion models \cite{sargent2023zeronvs,gao2024cat3d} and video diffusion models \cite{svd,voleti2024sv3d,chen2024v3d} to 3D scene generation. 
To extend the diffusion prior to complex, dynamic, large-scale driving scenes for 3D reconstruction, methods such as SGD \cite{sgd}, GGS \cite{han2024ggs} and MagicDrive3D \cite{magicdrive3d} employ generative models to broaden the range of training viewpoints. Nonetheless, these approaches mainly address sparse image data or static background elements, lacking the capacity to fully capture the complexities inherent in the 4D driving environments.

\begin{figure*}[!t]
\centering
\includegraphics[width=\textwidth]{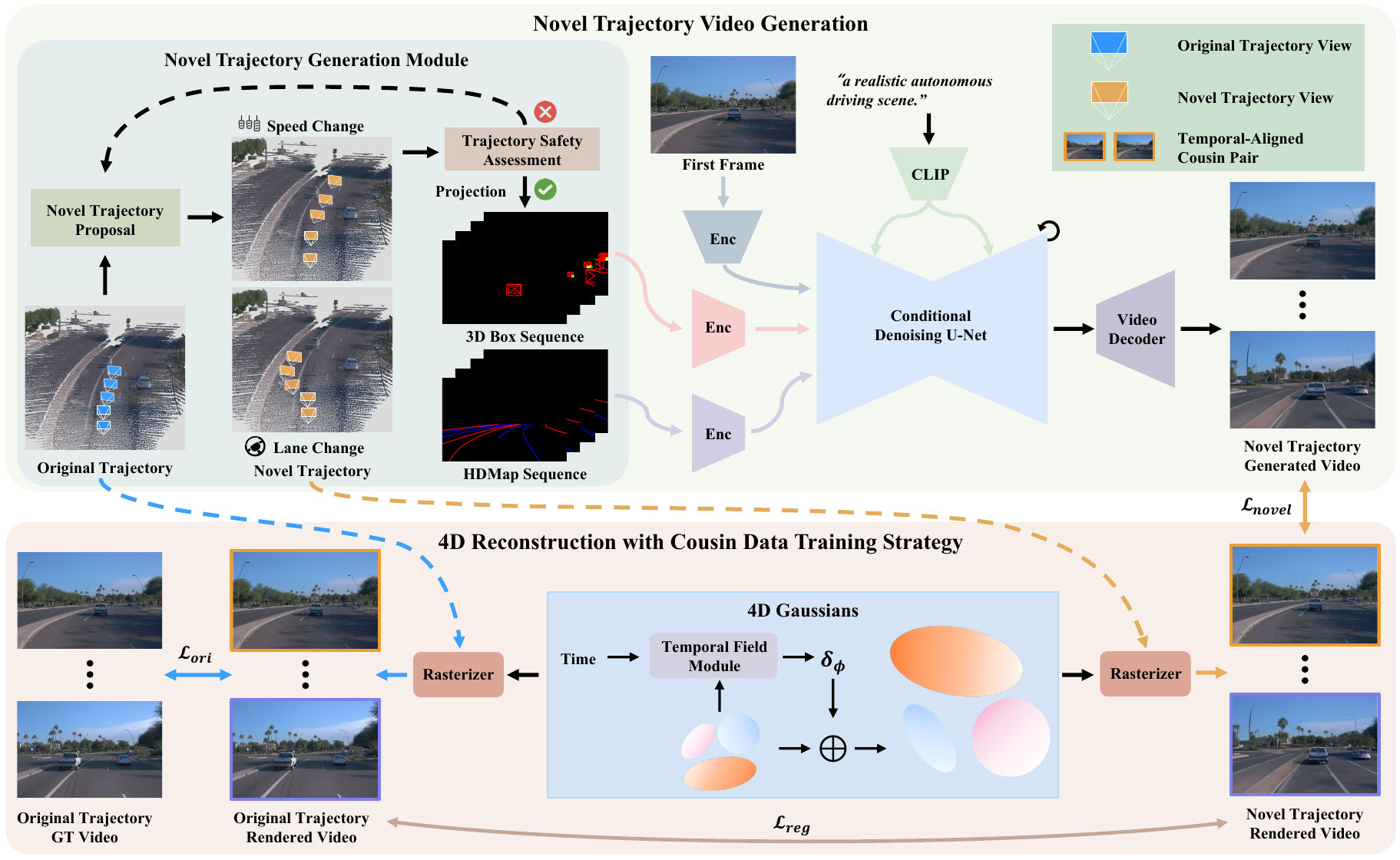}
\caption{The overall framework of \textit{DriveDreamer4D}. Initially, by altering the actions of the original trajectory (e.g., steering angle, speed), new trajectories can be obtained. Conditioned on the first frame and the structured information (3D bounding boxes, HDMap) from the new trajectory, the novel trajectory videos are generated. Subsequently, the temporal-aligned cousin pair (original and novel trajectory videos) are merged to optimize the 4D Gaussian Splatting model, where a regularization loss is calculated to ensure perceptual coherence.}
\label{fig_framework}
\vspace{-1em}
\end{figure*}

\section{Method}
In this section, we first elaborate on the preliminaries of 4D driving scene representation and world models for driving video generation. Then we present the details of \textit{DriveDreamer4D}, which enhances 4D driving scene representation leveraging priors from driving world models.
\subsection{Preliminary} 

\subsubsection{4D Driving Scene Representation}
% 4DGS represents the driving scene as a large number of 3DGS models and a temporal field module. Specifically, each 3DGS Models is parameterized as its center position $\boldsymbol{x}$, opacity $\boldsymbol{\gamma}$, covariance $\boldsymbol{\Sigma}$, and spherical harmonics parameters for view-dependent RGB color $\boldsymbol{c}$. Additionally, to make the optimization of 3D Gaussians more stable, the each covariance matrix $\boldsymbol{\Sigma}$ can be factorized into a scaling matrix $\boldsymbol{S}$ and a rotation matrix $\boldsymbol{R}$:
4DGS models the driving scene with a collection of 3DGS and a temporal field module. Each 3DGS \cite{3dgs} is parameterized by its center position $\boldsymbol{x}$, opacity $\boldsymbol{\gamma}$, covariance $\boldsymbol{\Sigma}$, and view-dependent RGB color $\boldsymbol{c}$, controlled via spherical harmonics. For stability, each covariance matrix $\boldsymbol{\Sigma}$ is decomposed by:
\begin{equation}
    {\boldsymbol{\Sigma}} = {\boldsymbol{RSS}}^{T}{\boldsymbol{R}}^{T},
\end{equation}
where scaling matrix $\boldsymbol{S}$ and a rotation matrix $\boldsymbol{R}$ are learnable parameters, represented by scaling $\boldsymbol{s}$ and quaternion $\boldsymbol{r}$. All trainable parameters of a single 3D Gaussian are collectively denoted as $\boldsymbol{\phi} = \{\boldsymbol{x, \gamma, s, r, c}\}$. The temporal field $\mathcal{F}$ takes $\boldsymbol{\phi}$ and a time step $\boldsymbol{t}_{gs}$ as input, outputting the offset $\delta \boldsymbol{\phi} = \{\delta \boldsymbol{x}, \delta \boldsymbol{\gamma}, \delta \boldsymbol{s}, \delta \boldsymbol{r}, \delta \boldsymbol{c}\}$ for each Gaussian relative to canonical space. The 4D Gaussian $\boldsymbol{\phi}' = \{\boldsymbol{x}', \boldsymbol{\gamma}', \boldsymbol{s}', \boldsymbol{r}', \boldsymbol{c}'\}$ is then computed by:

% The scaling matrix $\boldsymbol{S}$ and rotation matrix $\boldsymbol{R}$ can be represented the learnable parameters, scaling $\boldsymbol{s}$ and quaternion $\boldsymbol{r}$, respectively. For simplicity, we denote $\boldsymbol{\phi} = \{\boldsymbol{x,\gamma,s,r,c}\}$ as the set of all trainable parameters of a single 3D Gaussian Model. The temporal field module $\mathcal{F}$ takes the parameter $\boldsymbol{\phi}$ of each Gaussian and the current time step $\boldsymbol{t}_{gs}$ as input, producing the offset $\delta \boldsymbol{\phi}=\{\delta\boldsymbol{x},\delta\boldsymbol{\gamma},\delta\boldsymbol{s},\delta\boldsymbol{r},\delta\boldsymbol{c}\}$ of the Gaussian relative to canonical space. A 4D Gaussian $\boldsymbol{\phi}'=\{\boldsymbol{x}',\boldsymbol{\gamma}',\boldsymbol{s}',\boldsymbol{r}',\boldsymbol{c}'\}$ can be obtained by:
\begin{equation}
    \begin{aligned}
        {\boldsymbol{\phi}}' = {\boldsymbol{\phi}}+\delta{\boldsymbol{\phi}}
        ={\boldsymbol{\phi}}+\mathcal{F}({\boldsymbol{\phi},{\boldsymbol{t}_{gs}}}).
    \end{aligned}    
\end{equation}
Following \cite{yifan2019differentiable}, a differentiable Gaussian Splatting renderer is employed to project 4D Gaussians $\boldsymbol{\phi}$ into camera coordinates, yielding the covariance matrix ${\boldsymbol{\Sigma}'} = {\boldsymbol{JV\Sigma V}}^{T}{\boldsymbol{J}}^{T}$, where ${\boldsymbol{J}}$ is the Jacobian matrix of the perspective projection, and ${\boldsymbol{V}}$ is the transform matrix. 
The color of each pixel is calculated by $N$ ordered points using $\alpha$-blending:
\begin{equation}
    C = \sum_{i\in N}T_i{\boldsymbol{c}'_i}{\alpha_i},
\end{equation}
where $T_i$ is the transmittance defined by $\prod_{j=1}^{i-1}(1-{\alpha_j})$, $\boldsymbol{c}_i'$ denotes the color of each point, $\alpha_i$ is given by evaluating a
2D Gaussian with covariance $\boldsymbol{\Sigma}'$ multiplied with a learned per-point opacity $\boldsymbol{\gamma}_i'$.
The trainable parameters $\boldsymbol{\phi}'$ can be optimized by a combination of RGB loss, depth loss and \textrm{SSIM} loss:
\begin{equation}
\label{loss_ori}
\begin{aligned}
 \hspace{-4mm}
    \mathcal{L}_\text{ori}(\phi') &=\lambda_1\Vert \hat{I}_\text{ori}-I_\text{ori}\Vert_1+\lambda_2\Vert \hat{D}_\text{ori}-D_\text{ori}\Vert_1\\
    &+\lambda_3\textrm{SSIM}(\hat{I}_\text{ori},I_\text{ori}),
\end{aligned}
\end{equation}
where $\hat{I}_\text{ori}$ and $I_\text{ori}$ represent the rendered image and the ground truth image. $\hat{D}_\text{ori}$ and $D_\text{ori}$ are the rendered depth and the ground truth LiDAR depth map. $\textrm{SSIM}(\cdot)$ refers to the operation of the Structural Similarity Index Measure, and $\lambda_1,\lambda_2,\lambda_3$ are the loss weights.

\subsubsection{World Models for Driving Video Generation}

The world model module predicts possible future world states based on imagined action sequences \cite{lecun2022jepa}. 
Autonomous-driving world models \cite{drivedreamer,drivedreamer2,drivewm,vista}, typically based on diffusion models, leverage structured driving information or action controls to guide future video prediction. During training, these models first encode videos $\boldsymbol{v}$ into a lower-dimensional latent space $\boldsymbol{z} = \mathcal{E}(\boldsymbol{v})$ using a variational encoder $\mathcal{E}$. After adding noise $\epsilon_t$ to the latent, the diffusion model learns a denoising process. This diffusion process is optimized by:
\begin{equation}
\label{loss_diff}
    \begin{aligned}
        \mathcal{L}_{diff} = \mathbb{E}_{\boldsymbol{z}, \epsilon \sim \mathcal{N}(0,1), t}\left[\left\|\epsilon_t-\epsilon_{\theta}\left(\boldsymbol{z}_{t}, t,\boldsymbol{f}\right)\right\|_{2}^{2}\right],
    \end{aligned}
\end{equation}
where $\epsilon_{\theta}$ is a parameterized denoising network, $t$ denotes the time step, representing the level of noise added or removed at each stage. Additionally, to improve the controllability of the generated data, conditional features $\boldsymbol{f}$ (e.g., reference images, speed, steering angle, scene layouts, camera poses and textual information) can be introduced into the reverse diffusion process, ensuring that the generated outputs adhere to the input control signals. During inference, the world models can be conditioned on a reference image to control the style of the output scene, while predicting the future world states contingent upon the other input actions.

\subsection{DriveDreamer4D}
The overall pipeline of \textit{DriveDreamer4D} is depicted in Fig.~\ref{fig_framework}. In the upper part, the Novel Trajectory Generation Module (NTGM) is proposed to adjust driving actions (e.g., steering angle, speed) to generate new trajectories. These novel trajectories provide new perspectives for extracting structured information like 3D boxes and HDMap. Subsequently, a controllable video diffusion model synthesizes videos from these updated viewpoints, incorporating specific priors associated with the modified trajectories. In the lower part, the Cousin Data Training Strategy (CDTS) is introduced to combine the temporal-aligned original and generated data for optimizing the 4DGS model, where a regularization loss is calculated to impose perceptual coherence.
In the following sections, we delve into the details of novel trajectories video generation and then introduce the CDTS for 4D reconstruction.

\subsubsection{Novel Trajectory Video Generation}
\label{sec_3.2.1}
As previously mentioned, traditional 4DGS methods are limited in rendering complex maneuvers, largely due to the training data being dominated by straightforward driving scenarios. To overcome this, \textit{DriveDreamer4D} leverages world model priors to generate diverse viewpoint data, enhancing the 4D scene representation. To achieve this, we propose the NTGM, which is designed to create new trajectories that serve as input for the world model, enabling the automated generation of complex maneuver data. NTGM comprises two main components: (1) novel trajectory proposal, (2) trajectory safety assessment. In the novel trajectory proposal stage, \textit{text-to-trajectory} \cite{drivedreamer2} can be adopted to automatically generate diverse complex trajectories. Additionally, trajectories can be custom-designed to meet specific requirements, allowing for tailored data generation based on precise needs. 
The overview of the custom-designed trajectory proposal (e.g., lane change) and trajectory safety assessment is shown in the Algo.~\ref{alg_ntgm}. In a specific driving scenario, the original trajectory in the world coordinate system can be readily acquired as $\mathcal{T}_\text{ori}^\text{world}=\{p_i^\text{world}\}_{i=0}^{K}$, where $K$ denotes the number of frames and $p_i^\text{world}\in \mathbb{R}^{3}$ refers to the position of the ego-vehicle at the $i$-th frame. To propose novel trajectories, the original trajectory $\mathcal{T}_\text{ori}^\text{world}$ is transformed into the ego-vehicle coordinate system of the first frame, denoted as $\mathcal{T}_\text{ori}^\text{EgoStart}$ and computed as:
\begin{equation}
    \begin{aligned}
        [p_i^\text{EgoStart}, 1]^T = M_0^{-1}\times [p_i^\text{world},1]^T,
    \end{aligned}
\end{equation}
where $M_0 \in \mathbb{R}^{4\times 4}$ represents the transformation matrix from the ego-vehicle coordinate system of the first frame to the world coordinate system, $[\cdot]$ denotes the operation of the concat. In the ego-vehicle coordinate system, the vehicle’s heading is aligned with the positive $x$-axis, the $y$-axis points to the left side of the vehicle, and the $z$-axis is oriented vertically upwards, perpendicular to the plane of the vehicle. Consequently, changes in the vehicle’s velocity and direction can be respectively represented by adjusting the value along the $x$-axis and $y$-axis. A final safety assessment is conducted for the newly generated trajectory points, which includes verifying whether the vehicle trajectories $p$ remain within drivable areas $\mathcal{B}_{\text{road}}$ and ensuring that no collisions occur with pedestrians or other vehicles $\{o_j\}_{j=1}^M$.
% \begin{equation}
%      p_i \in \mathcal{B}_{\text{road}}, 
% \end{equation}
% \begin{equation}
%      \| p_i - o_j \| \geq d_{\text{min}}, \quad \forall j \in \{1, \dots, M\}.
% \end{equation}
\begin{equation}
\begin{array}{ll}
    p \in \mathcal{B}_{\text{road}}, & \\[8pt]
    \| p - o_j \| \geq d_{\text{min}}, & \quad \forall j \in \{1, \dots, M\},
\end{array}
\end{equation}
where $d_{\text{min}}$ is the minimal distance between different agents.
Once a novel trajectory that complies with traffic regulations is generated, the road structure and 3D bounding boxes can be projected onto the camera view from the perspective of the new trajectory, thereby generating structured information relative to the updated trajectory. This structured information, along with the initial frame and text, is fed into a world model \cite{drivedreamer2} to produce the videos that follow the novel trajectories.

\begin{algorithm}
    \caption{Novel Trajectory Generation Module}
    \label{alg_ntgm}
    \renewcommand{\algorithmicrequire}{\textbf{Input:}}
    \renewcommand{\algorithmicensure}{\textbf{Output:}}
    \begin{algorithmic}
        \REQUIRE Trajectory $\mathcal{T}_{\text{ori}}^{\text{world}}$, Transformation matrix $M_0$
        % , type of novel trajectory $type$
        \ENSURE Novel trajectory $\mathcal{T}_{\text{novel}}^{\text{ego}}$
        \STATE $\mathcal{T}_{\text{novel}}^{\text{ego}}\gets [[0, 0, 0]]$ 
        \STATE Offset $\gets$ 0
        \FOR{each $p^{\text{world}}_\text{{ori}}$ in $\mathcal{T}_{\text{ori}}^{\text{world}}$[1:]}
        % \vspace{0.3em}
        \STATE $p^{\text{Ego}\text{Start}} \gets$  RelativeCoord($p^{\text{world}}_\text{{ori}}$, $M_0$)
        \STATE MaxOffset $\gets$ 0.1
        \WHILE{True}
           \STATE NewOffset $\gets$ Offset + RandOffset(0, MaxOffset)
           \STATE $p^{\text{Ego}\text{Start}'}\gets p^{\text{Ego}\text{Start}}$ + [0, NewOffset, 0]
            % \IF{$type$==``accelerate"}
            %     \STATE $new\_p^{ego\_start}\gets p^{ego\_start}$+[new\_offset,0,0] 
            % \ELSIF{$type$==``decelerate"}
            %     \STATE $new\_p^{ego\_start}\gets p^{ego\_start}$-[new\_offset,0,0]  
            % \ELSIF{$type$==``left\_change"}
            %     \STATE $new\_p^{ego\_start}\gets p^{ego\_start}$+[0,new\_offset,0] 
            % \ELSIF{$type$==``right\_change"}
            %     \STATE $new\_p^{ego\_start}\gets p^{ego\_start}$-[0,new\_offset,0]
            % \ENDIF
            \IF{SafeCheck($p^{\text{EgoStart}'}$)} 
                \STATE AddElement($\mathcal{T}_{\text{novel}}^{\text{ego}}$, $p^{\text{EgoStart}'}$)
                \STATE Offset $\gets$ NewOffset
                \STATE \textbf{break}
            \ELSE
                \STATE MaxOffset $\gets$ MaxOffset/2
            \ENDIF
        \ENDWHILE
        \ENDFOR
    
    \end{algorithmic}
\end{algorithm}

%所有新视角的指标
\begin{table*}[t]
\centering
\setlength{\abovecaptionskip}{0.5em}
\resizebox{1\linewidth}{!}{
\begin{tabular}{@{}lccccccccc@{}}
\toprule
\multirow{2}{*}{Method} & \multicolumn{2}{c}{Lane Change} & \multicolumn{2}{c}{Acceleration} & \multicolumn{2}{c}{Deceleration} &\multicolumn{2}{c}{Average} \\ \cmidrule(lr){2-3} \cmidrule(lr){4-5} \cmidrule(lr){6-7} \cmidrule(lr){8-9}
 & NTA-IoU $\uparrow$ & NTL-IoU $\uparrow$ & NTA-IoU $\uparrow$ & NTL-IoU $\uparrow$ & NTA-IoU $\uparrow$ & NTL-IoU $\uparrow$ & NTA-IoU $\uparrow$ & NTL-IoU $\uparrow$ \\ \midrule
PVG \cite{pvg} & 0.256 & 50.70 & 0.396 & 53.08 & 0.394 & 53.65 & 0.349 & 52.48\\
\textit{DriveDreamer4D} with PVG & \textbf{0.438} & \textbf{53.06} & \textbf{0.421} & \textbf{53.35} & \textbf{0.424} & \textbf{53.89} & \textbf{0.428} & \textbf{53.43}\\
\midrule
$\text{S}^3$Gaussian \cite{s3gaussian} & 0.175 & 49.05 & 0.434 & 51.93 & 0.384 & 52.14&  0.331&  51.04\\
\textit{DriveDreamer4D} with $\text{S}^3$Gaussian & \textbf{0.495} & \textbf{53.42} & \textbf{0.484} & \textbf{52.63} & \textbf{0.445} & \textbf{52.69} & \textbf{0.475} &  \textbf{52.91}\\
\midrule
Deformable-GS \cite{deformablegs} & 0.240 & 51.62 & 0.346 & 52.17 & 0.377 & 53.21 & 0.321 & 52.33\\
\textit{DriveDreamer4D} with Deformable-GS & \textbf{0.335} & \textbf{52.93} & \textbf{0.371} & \textbf{52.77} & \textbf{0.406} & \textbf{53.79} & \textbf{0.371} & \textbf{53.16}\\
\bottomrule
\end{tabular}}
\caption{Comparison of NTA-IoU and NTL-IoU scores across different novel trajectory views (lane change, acceleration, deceleration).}
\label{tab:method_comparison}
\vspace{-1em}
\end{table*}

% \subsubsection{4D Reconstruction with Video Diffusion Priors}
\subsubsection{Cousin Data Training Strategy}

To better integrate generated data for training 4DGS, we propose the CDTS. Specifically, we construct temporally aligned cousin pair data as a minimal training batch:
\begin{equation}
    \text{BatchStack}(\{\hat{I}_{\text{ori},t}\}_{t=0}^{T}, \{\hat{I}_{\text{novel},t}\}_{t=0}^{T}),
\end{equation}
where $\text{BatchStack}(\cdot)$ is the data processor to stack temporal-aligned $\{\hat{I}_{\text{ori},t}\}_{t=0}^{T}$ and $\{\hat{I}_{\text{novel},t}\}_{t=0}^{T}$ into a training batch.
By leveraging real and synthetic data aligned at each timestep, CDTS mitigates the data gap in 4DGS training, enhancing the model’s ability to learn consistent representations across real and synthetic data.
To optimize 4DGS, the temporal-aligned cousin pair is input per step before gradient optimization. The loss function $\mathcal{L}_\text{ori}$ for the original data is defined in Eq. \ref{loss_ori}. And the loss function $\mathcal{L}_\text{novel}$ for the generated data is akin to \cite{pvg,s3gaussian}, defined as follows:
\begin{equation}
\begin{aligned}
 % \hspace{-2mm}
    \mathcal{L}_\text{novel}(\phi')&=\lambda_\text{1}\Vert \hat{I}_\text{novel}-I_\text{novel}\Vert_1 \\
    &\quad+\lambda_\text{3}\textrm{SSIM}(\hat{I}_\text{novel},I_\text{novel}),
\end{aligned}
\end{equation}
where $I_\text{novel}$ represents the generated images corresponding to the novel trajectories as described in Sec. \ref{sec_3.2.1}, and $\hat{I}_\text{novel}$ denotes the rendered images under the novel trajectories via differentiable splatting \cite{yifan2019differentiable}. Notably, different from \cite{pvg,s3gaussian}, depth maps are not employed as constraints in the optimization of 4DGS when using the generated dataset $D_\text{novel}$. The limitation arises from the fact that LiDAR point cloud data is exclusively collected for the original trajectory. When these LiDAR points are projected onto a new trajectory, it cannot produce a complete depth map for the new perspective, as something visible in the novel trajectory may have been occluded in the original view. Consequently, the incorporation of such depth maps does not facilitate the optimization of the 4DGS model. More details are described in Sec. \ref{sec_4.3}. Additionally, we propose a regularization loss to enhance the perceptual coherence, defined as follows:
\begin{equation}
    \mathcal{L}_\text{reg}(\phi')=\Vert \mathcal{F}_p(\hat{I}_\text{ori})-\mathcal{F}_p(\hat{I}_\text{novel})\Vert_1,
\end{equation}
where $\mathcal{F}_p(\cdot)$ denotes the perception feature extraction model \cite{fid}. The overall loss function for mixed training is defined as follows:
\begin{equation}
    \mathcal{L}(\phi')=\mathcal{L}_\text{ori}+\lambda_\text{novel}\mathcal{L}_\text{novel}+\lambda_{\text{reg}}\mathcal{L}_\text{reg}.
\end{equation}

\section{Experiments}
In this section, we first outline the experimental setup, including the dataset, implementation details, and evaluation metrics. Subsequently, both quantitative and qualitative evidence are provided to demonstrate that the proposed \textit{DriveDreamer4D} significantly enhances rendering quality for novel trajectory viewpoints and improves the spatiotemporal coherence of foreground and background components. Finally, we conduct an ablation study on the hyperparameter settings, as well as the effects of depth loss and the proposed CDTS, including temporal-aligned pairs, and regularization loss.

\subsection{Experiment Setup}

\noindent
\textbf{Dataset.}
We conduct experiments using the Waymo dataset \cite{waymo}, known for its comprehensive real-world driving logs. However, most logs capture scenes with relatively straightforward dynamics, lacking focus on scenarios with dense, complex vehicle interactions. To address this gap, we specifically select eight scenes characterized by highly dynamic interactions, featuring numerous vehicles with diverse relative positions and intricate driving trajectories. Each selected segment contains approximately 40 frames, with segment IDs detailed in the supplement.

\noindent
\textbf{Implementation Details.}
To demonstrate the versatility and robustness of \textit{DriveDreamer4D}, we incorporate various 4DGS baselines into our pipeline, including Deformable-GS \cite{deformablegs}, $\text{S}^3$Gaussian \cite{s3gaussian}, and PVG \cite{pvg}. For a fair comparison, LiDAR supervision is introduced to Deformable-GS. During training, scenes are segmented into multiple clips, each containing 40 frames, aligned with the generative model's output length. We use only forward-facing camera data and standardize the resolution across methods to $640\times 960$. Our models are trained using the Adam optimizer \cite{kingma2014adam}, following the learning rate schedule used for 3D Gaussian Splatting \cite{3dgs}. Hyperparameter settings are aligned with each baseline \cite{deformablegs,s3gaussian,pvg}, and the training strategy remains the same, with the exception of the incorporation of CDTS.

% with 50,000 iterations for \cite{deformablegs,s3gaussian} and 30,000 iterations for \cite{pvg}.
% We aim to compare these advanced techniques with foundational algorithms, ensuring fair comparisons by implementing LiDAR supervision for Deformable-GS and PVG and adhering to official implementations for the others. 

% EmerNeRF leads the field in reconstructing dynamic driving environments using a 3D Hash-Grid for static components and a 4D Hash-Grid for dynamic elements, enhanced with a flow field to effectively capture features from dynamic activities. Deformable-GS, in its approach, conceptualizes a canonical space with Gaussian representations, employing a deformation network to dynamically adjust Gaussian properties to scene changes. PVG introduces Periodic Vibration Gaussians that oscillate with tunable parameters such as vibration direction, lifespan, and peak opacity, optimized via self-supervision for effective decomposition of dynamic scenes. Similarly, $\text{S}^3$Gaussian, a novel self-supervised algorithm, excels in decomposing both dynamic and static 3D Gaussians in urban street scenes, using a sophisticated spatial-temporal decomposition network to process complex dynamics without the need for manually annotated data. 

% ##########################################
%变道新视角
\begin{figure*}[!t]
\centering
\setlength{\abovecaptionskip}{0em}
\includegraphics[width=0.94\textwidth]{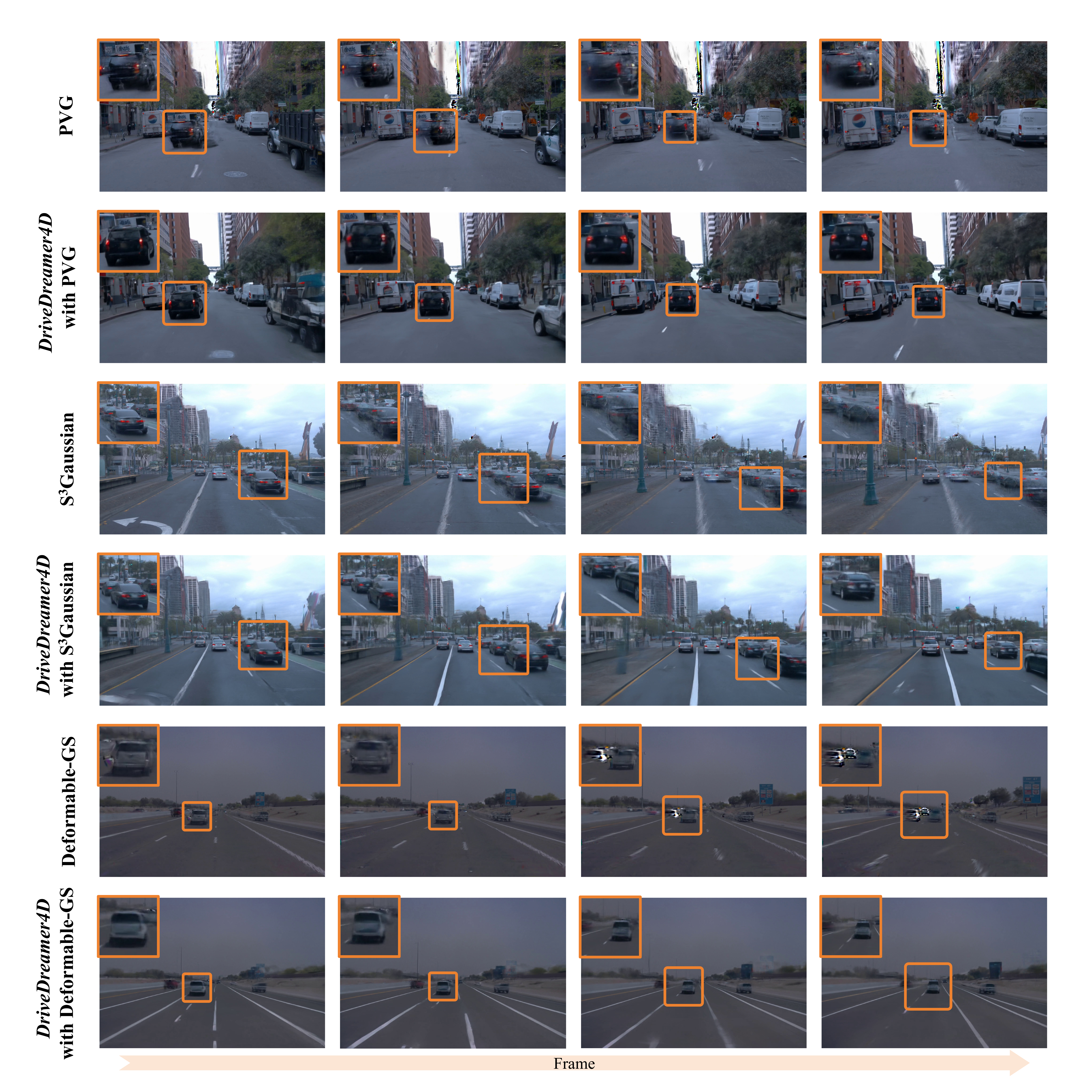}
\caption{Qualitative comparisons of novel trajectory renderings during lane change scenarios. The orange boxes highlight that \textit{DriveDreamer4D} significantly enhances the rendering quality across various baselines (PVG \cite{pvg}, $\text{S}^3$Gaussian \cite{s3gaussian}, Deformable-GS \cite{deformablegs}).}
\label{fig:change}
\vspace{-1em}
\end{figure*}
% ##########################################

\noindent
\textbf{Metrics.} 
% Unlike traditional 3D reconstruction tasks, assessing 4D reconstruction in dynamic driving scenarios requires evaluating not only the rendering quality of static elements but also the spatiotemporal coherence of various traffic components. Traditional metrics such as PSNR and SSIM primarily measure pixel-level quality and structural similarity, overlooking the temporal and spatial consistency critical for dynamic agents in driving contexts.
Traditional 3D reconstruction tasks typically employ PSNR and SSIM metrics for evaluation, with validation sets that closely match the training data distribution (i.e., uniformly sampling frames from video sequences for validation, with the remainder used for training). However, in closed-loop driving simulation, the focus shifts to evaluating model rendering performance under novel trajectories, where corresponding sensor data are unavailable, making metrics like PSNR and SSIM inapplicable for evaluation.
Therefore, we propose Novel Trajectory Agent IoU (NTA-IoU) and Novel Trajectory Lane IoU (NTL-IoU), which assess the spatiotemporal coherence of foreground and background traffic components in novel trajectory viewpoints. 
%这个是变道的fid比较
\begin{table}[t]
\centering
\setlength{\abovecaptionskip}{0.5em}
\begin{tabular}{@{}l|c@{}}
\toprule
Method & FID $\downarrow$ \\ \midrule
PVG\cite{pvg} & 105.29 \\
\textit{DriveDreamer4D} with PVG & \textbf{71.52} \\
\midrule
$\text{S}^3$Gaussian\cite{s3gaussian} & 124.90\\
\textit{DriveDreamer4D} with $\text{S}^3$Gaussian & \textbf{66.93}\\
\midrule
Deformable-GS\cite{deformablegs} & 92.34\\
\textit{DriveDreamer4D} with Deformable-GS & \textbf{77.32}\\
\bottomrule
\end{tabular}
\caption{Comparison of FID scores in novel trajectory view synthesis (lane change) on the Waymo dataset.}
\label{tab:fid}
\vspace{-1em}
\end{table}

For NTA-IoU, we use YOLO11 \cite{yolo11_ultralytics} to identify vehicles in images rendered from novel trajectory views, yielding 2D bounding boxes. Simultaneously, geometric transformations are applied to the original 3D bounding boxes, projecting them onto the new viewpoints to generate corresponding 2D bounding boxes. For each projected 2D box, we then identify the closest detector-generated 2D box and compute their Intersection over Union (IoU). 
To ensure accurate matching, a distance threshold $d_{\text{thresh}}$ is introduced: when the center-to-center distance $\|c(B^{\text{proj}}) - c(B^{\text{det}})\|$ between the nearest detected box $B^{\text{det}}$ and the correctly projected box $B^{\text{proj}}$ surpasses this threshold, their NTA-IoU is assigned a value of zero:
\begin{equation}
\hspace{-0.2em}
    \text{NTA-IoU} = 
    \begin{cases} 
      0 \quad\text{if } \|c(B^{\text{proj}}) - c(B^{\text{det}})\| \ge d_{\text{thresh}} & \\
      \text{IoU}(B^{\text{proj}}, B^{\text{det}})  \quad\quad\quad\quad \text{otherwise}. &
    \end{cases} 
\hspace{-0.6em}
\end{equation}
For NTL-IoU, we employ TwinLiteNet \cite{che2023twinlitenet} to extract 2D lanes from rendered images. Ground truth lanes are also projected onto the 2D image plane. We then compute the mean Intersection-over-Union (mIoU) between the rendered and ground truth lanes $L^{\text{det}}$ and $L^{\text{proj}}$:
\begin{equation}
    \text{NTL-IoU} = \text{mIoU}(L^{\text{proj}}, L^{\text{det}}).
\end{equation}

%##################################################
% \begin{figure*}[!ht] %变速新视角
% \centering
% % \setlength{\abovecaptionskip}{0.2em}
% \includegraphics[width=\textwidth]{figures/speed_change.pdf}
% \caption{Qualitative comparisons of novel trajectory renderings during speed change scenarios. The orange boxes highlight that \textit{DriveDreamer4D} significantly enhances the rendering quality across various baseline methods (PVG \cite{pvg}, $\text{S}^3$Gaussian \cite{s3gaussian}, Deformable-GS \cite{deformablegs}).}
% \label{fig:speed}
% %\vspace{-1.5em}
% \end{figure*}
\noindent
Additionally, in lane change scenarios, we observe inaccuracies in relative positioning, as well as frequent occurrences of artifacts such as flying points and ghosting, which notably degrade image quality. To assess this, we employ the FID metric \cite{fid}, which quantifies differences in feature distribution between rendered novel trajectory images and original trajectory images. This metric effectively reflects visual quality and is particularly sensitive to artifacts like flying points and ghosting, providing a robust measure of image fidelity in these complex scenes. Finally, a user study is conducted to evaluate the quality of the renderings, where participants compare the rendering results of each baseline with its \textit{DriveDreamer4D} enhanced version across three novel trajectories. The evaluation criteria focus on overall video quality, with particular attention to foreground objects like vehicles. For each comparison, participants were asked to select the option they found most favorable. Further details are provided in the supplement.

% When reconstructing four-dimensional autonomous driving scenes using 4DGS, rendering new viewpoints for actions such as ego-vehicle acceleration, lane change, or deceleration can lead to inaccuracies in the relative positioning of surrounding vehicles. This issue primarily stems from the sparsity of the image sets used for training. Due to this limitation, 4DGS is unable to accurately guide the temporal changes of the Gaussian sphere. This problem is particularly pronounced during lane change maneuvers; for example, when generating a new viewpoint for a lane change, the surrounding vehicles may incorrectly move left or right along with the ego-vehicle.

% Moreover, current metrics like PSNR and SSIM do not intuitively assess the accuracy of inter-vehicle relationships in dynamic scenes such as lane change or overtaking. These traditional indices focus primarily on pixel-level image quality and structural similarity. However, in the reconstruction of 4D scenes, the relative interactions between moving objects become more crucial. Additionally, due to limitations in data collection, new viewpoints often lack actual images to evaluate rendering quality, complicating the assessment of these dynamic scenes.

% Moreover, due to limitations in sensor data collection, conventional evaluation methods often lack real-world images, making it challenging to comprehensively assess rendering quality.

\begin{table}[t]
\centering
\setlength{\abovecaptionskip}{0.4em}
\resizebox{1\linewidth}{!}{
\begin{tabular}{l|cccc}
\toprule
\multirow{2}{*}{Counterpart Method} & \multicolumn{4}{c}{\textit{DriveDreamer4D} Win Rate} \\& Lane Change & Acceleration & Deceleration  & Average
\\ \midrule
PVG \cite{pvg} & 100.0\% & 90.5\%  & 89.1\% & \textbf{93.2\%} \\

$\text{S}^3$Gaussian \cite{s3gaussian} & 100.0\% & 97.9\%  & 92.2\% & \textbf{96.7\%} \\

Deformable-GS \cite{deformablegs} & 95.8\% & 83.5\%  & 72.9\% & \textbf{84.1\%} \\
\bottomrule
\end{tabular}}
\caption{User study comparison of \textit{DriveDreamer4D} win rates across various novel trajectory view synthesis.}
\label{tab:user}
\vspace{-1.5em}
\end{table}

\subsection{Comparison with Different 4DGS Baselines}
\label{sec4.2}

\noindent
\textbf{Quantitative Results.} As demonstrated in Tab.~\ref{tab:method_comparison}, integrating \textit{DriveDreamer4D} with different 4DGS algorithms consistently yields superior NTA-IoU and NTL-IoU scores across diverse, complex maneuvers (e.g., lane changes, acceleration, and deceleration), significantly outperforming the baseline methods. Specifically, with \textit{DriveDreamer4D}, the average NTA-IoU scores across three baselines (PVG \cite{pvg}, $\text{S}^3$Gaussian \cite{s3gaussian}, Deformable-GS \cite{deformablegs}) are relatively enhanced by 22.6\%, 43.5\%, and 15.6\%, underscoring \textit{DriveDreamer4D}’s capability to improve the spatiotemporal coherence of foreground agents. Moreover \textit{DriveDreamer4D} facilitates a relative improvement in average NTL-IoU for these baselines by 1.8\%, 3.7\%, and 1.6\%, thereby markedly enhancing the spatiotemporal coherence of background lanes in 4D rendering of driving scenarios.

In addition to verifying the spatiotemporal consistency of rendered novel trajectory views, we leverage the FID metric to assess rendering quality under novel trajectories. Given that acceleration and deceleration scenarios yield rendered views with distributional similarities to ground truth, limiting FID's discriminative capability across algorithms, our FID comparisons focus specifically on lane change scenarios. Experiment results, as presented in Tab.~\ref{tab:fid}, indicate that our method substantially outperforms the baseline methods (PVG \cite{pvg}, $\text{S}^3$Gaussian \cite{s3gaussian}, Deformable-GS \cite{deformablegs}), with FID relative improvements of 32.1\%, 46.4\%, and 16.3\%. These results highlight \textit{DriveDreamer4D}’s capability to enhance generation quality for novel trajectory viewpoints.

Finally, we conduct a user study to evaluate the rendering quality of different methods on novel trajectories, with a specific focus on foreground agents. For each method, we generate three novel trajectory views—lane change, acceleration, and deceleration—across eight scenes from the Waymo dataset \cite{waymo}. Participants are then asked to select the renderings they found most visually favorable in each comparison. The \textit{DriveDreamer4D} win rates from this study, shown in Tab.~\ref{tab:user}, reveal a significant user preference for our method’s renderings.

\begin{table}[t]
    \centering
        
    \begin{minipage}[t]{0.23\textwidth}
        % \makeatletter\def\@captype{table}
        \centering
        \setlength{\abovecaptionskip}{0.4em}
        \begin{tabular}{m{2.2em}m{2.5em}m{2.8em}}
        \toprule
        $\lambda_\text{novel}$ & NTA-IoU $\uparrow$  & FID $\downarrow$\\ \midrule
        0 & 0.349 & 105.29  \\
        % \midrule
        0.5 & 0.405 & 82.84\\
        1 & \textbf{0.420} & \textbf{79.54}\\
        1.5 & 0.417 & 82.10\\
        \bottomrule
        \end{tabular}
        \caption{Ablation study on the training loss weight $\lambda_\text{novel}$ for novel trajectory data.}
        \label{tab:loss_novel}
        % \end{table}
    \end{minipage}
    \hspace{0.5em}
    \begin{minipage}[t]{0.22\textwidth}
    % \makeatletter\def\@captype{table}
        % \begin{table}[t]
        \centering
        \setlength{\abovecaptionskip}{0.4em}
        \begin{tabular}{m{2em}m{2.5em}m{2.5em}}
        \toprule
        $\lambda_\text{reg}$ & NTA-IoU $\uparrow$ & FID $\downarrow$\\ \midrule
        0 & 0.420 & 79.54 \\
        % \midrule
        1e-2 & 0.411 & 119.39 \\
        1e-3 & \textbf{0.428} & \textbf{71.52} \\
        1e-4 & 0.422 & 75.31 \\
        \bottomrule
        \end{tabular}
        \caption{Ablation study on the regularization loss weight $\lambda_\text{reg}$.}
        \label{tab:loss_reg}
        % \end{table}
    \end{minipage}
% \vspace{-1em}
\end{table}

\begin{table}[t]
\centering
\setlength{\abovecaptionskip}{0.4em}
\resizebox{1\linewidth}{!}{
\begin{tabular}{cccccc}
\toprule
\makecell[c]{Depth\\Loss} & \makecell[c]{Temporal-aligned\\Cousin Pair} & \makecell[c]{Regularization\\Loss} & \makecell[c]{NTA-IoU} $\uparrow$ & FID $\downarrow$\\ \midrule
$\checkmark$ & $\times$ &$\times$& 0.401 & 82.63\\
$\times$ & $\times$ & $\times$ & 0.420 & 79.54\\
$\times$ & $\checkmark$ & $\times$ & 0.423 & 76.20\\
$\times$ & $\checkmark$ & $\checkmark$ & \textbf{0.428} & \textbf{71.52}\\
\bottomrule
\end{tabular}}
\caption{Ablation study on the depth loss and CDTS.}
\label{tab:ablation}
\vspace{-1.5em}
\end{table}

\noindent
\textbf{Qualitative Results.}
In addition to quantitative comparisons, we provide a qualitative analysis of novel trajectory view renderings. As shown in Fig.~\ref{fig:change}, we present the novel trajectory view synthesis during lane change.  Images rendered by the baseline algorithms exhibit issues where foreground vehicles incorrectly change lanes in sync with the camera's motion, and some vehicles are incompletely rendered.  Additionally, the background is filled with speckles and ghosting. Especially shown in the rightmost column of Fig.~\ref{fig:change}, baseline algorithms often produce blurred, ghosted foreground vehicles and background speckles in the sky, alongside blurred lane markings. Our method, however, significantly improves rendering quality, as highlighted by the orange boxes. Vehicle contours are sharper, and background artifacts such as speckles and ghosting are substantially reduced.

\subsection{Ablation Studies}
\label{sec_4.3}
We conduct an ablation study based on PVG \cite{pvg}, analyzing the effects of hyperparameters settings for $\lambda_\text{novel}$ and $\lambda_\text{reg}$, as well as the impact of depth loss, temporal-aligned cousin pairs, and regularization loss. Following the experiments shown in Tab.~\ref{tab:loss_novel} and~\ref{tab:loss_reg}, we set $\lambda_\text{novel}$ to 1 and $\lambda_\text{reg}$ to $1\times10^{-3}$, which yield the best performance. As shown in Tab.~\ref{tab:ablation}, the experiment confirms that the depth loss should be excluded when optimizing novel trajectory views, since LiDAR depth maps are incomplete due to occlusions. Furthermore, employing temporal-aligned cousin pairs achieves an FID of 76.20 and an NTA-IoU of 0.423, with the perception regularization loss further improving the FID to 71.52 and NTA-IoU to 0.428. These results highlight the effectiveness of CDTS, in achieving $\sim$10\% improvement in FID, along with a $\sim$2\% increase in NTA-IoU.
\section{Discussion and Conclusion}
% \vspace{-0.5em}
In this paper, we presented \textit{DriveDreamer4D}, a novel framework designed to advance 4D driving scene representations by harnessing priors from world models. Addressing key limitations of current sensor simulation methods—namely, their dependence on forward-driving training data distributions and inability to model complex maneuvers—\textit{DriveDreamer4D} leverages a world model to generate novel trajectory videos that complement real-world driving data. By explicitly employing structured conditions, our framework maintains spatial-temporal consistency across traffic elements, ensuring that generated data adheres closely to the dynamics of real-world traffic scenarios.
Our experiments demonstrate that \textit{DriveDreamer4D} achieves superior quality in generating diverse simulation viewpoints, with significant improvements in both the rendering fidelity and spatiotemporal coherence of scene components. Notably, these results highlight \textit{DriveDreamer4D}'s potential as a foundation for closed-loop simulations that require high-fidelity reconstructions of dynamic driving scenes.

{
    \small
    \bibliographystyle{ieeenat_fullname}
    \bibliography{PaperForReview}

\begin{thebibliography}{83}
\providecommand{\natexlab}[1]{#1}
\providecommand{\url}[1]{\texttt{#1}}
\expandafter\ifx\csname urlstyle\endcsname\relax
  \providecommand{\doi}[1]{doi: #1}\else
  \providecommand{\doi}{doi: \begingroup \urlstyle{rm}\Url}\fi

\bibitem[Attal et~al.(2023)Attal, Huang, Richardt, Zollhoefer, Kopf, O’Toole, and Kim]{attal2023hyperreel}
Benjamin Attal, Jia-Bin Huang, Christian Richardt, Michael Zollhoefer, Johannes Kopf, Matthew O’Toole, and Changil Kim.
\newblock Hyperreel: High-fidelity 6-dof video with ray-conditioned sampling.
\newblock In \emph{CVPR}, 2023.

\bibitem[Barron et~al.(2022)Barron, Mildenhall, Verbin, Srinivasan, and Hedman]{mipnerf}
Jonathan~T Barron, Ben Mildenhall, Dor Verbin, Pratul~P Srinivasan, and Peter Hedman.
\newblock Mip-nerf 360: Unbounded anti-aliased neural radiance fields.
\newblock In \emph{CVPR}, 2022.

\bibitem[Barron et~al.(2023)Barron, Mildenhall, Verbin, Srinivasan, and Hedman]{zipnerf}
Jonathan~T Barron, Ben Mildenhall, Dor Verbin, Pratul~P Srinivasan, and Peter Hedman.
\newblock Zip-nerf: Anti-aliased grid-based neural radiance fields.
\newblock In \emph{ICCV}, 2023.

\bibitem[Blattmann et~al.(2023{\natexlab{a}})Blattmann, Dockhorn, Kulal, Mendelevitch, Kilian, Lorenz, Levi, English, Voleti, Letts, et~al.]{svd}
Andreas Blattmann, Tim Dockhorn, Sumith Kulal, Daniel Mendelevitch, Maciej Kilian, Dominik Lorenz, Yam Levi, Zion English, Vikram Voleti, Adam Letts, et~al.
\newblock Stable video diffusion: Scaling latent video diffusion models to large datasets.
\newblock \emph{arXiv preprint arXiv:2311.15127}, 2023{\natexlab{a}}.

\bibitem[Blattmann et~al.(2023{\natexlab{b}})Blattmann, Rombach, Ling, Dockhorn, Kim, Fidler, and Kreis]{videoldm}
Andreas Blattmann, Robin Rombach, Huan Ling, Tim Dockhorn, Seung~Wook Kim, Sanja Fidler, and Karsten Kreis.
\newblock Align your latents: High-resolution video synthesis with latent diffusion models.
\newblock In \emph{CVPR}, 2023{\natexlab{b}}.

\bibitem[Brooks et~al.(2024)Brooks, Peebles, Holmes, DePue, Guo, Jing, Schnurr, Taylor, Luhman, Luhman, Ng, Wang, and Ramesh]{videoworldsimulators2024}
Tim Brooks, Bill Peebles, Connor Holmes, Will DePue, Yufei Guo, Li Jing, David Schnurr, Joe Taylor, Troy Luhman, Eric Luhman, Clarence Ng, Ricky Wang, and Aditya Ramesh.
\newblock Video generation models as world simulators.
\newblock 2024.

\bibitem[Caesar et~al.(2020)Caesar, Bankiti, Lang, Vora, Liong, Xu, Krishnan, Pan, Baldan, and Beijbom]{nusc}
Holger Caesar, Varun Bankiti, Alex~H Lang, Sourabh Vora, Venice~Erin Liong, Qiang Xu, Anush Krishnan, Yu Pan, Giancarlo Baldan, and Oscar Beijbom.
\newblock nuscenes: A multimodal dataset for autonomous driving.
\newblock In \emph{CVPR}, 2020.

\bibitem[Che et~al.(2023)Che, Nguyen, Pham, and Lam]{che2023twinlitenet}
Quang-Huy Che, Dinh-Phuc Nguyen, Minh-Quan Pham, and Duc-Khai Lam.
\newblock Twinlitenet: An efficient and lightweight model for driveable area and lane segmentation in self-driving cars.
\newblock In \emph{MAPR}, 2023.

\bibitem[Chen et~al.(2023)Chen, Gu, Jiang, Zhu, and Zhang]{pvg}
Yurui Chen, Chun Gu, Junzhe Jiang, Xiatian Zhu, and Li Zhang.
\newblock Periodic vibration gaussian: Dynamic urban scene reconstruction and real-time rendering.
\newblock \emph{arXiv preprint arXiv:2311.18561}, 2023.

\bibitem[Chen et~al.(2024{\natexlab{a}})Chen, Wang, Wang, Wang, and Liu]{chen2024v3d}
Zilong Chen, Yikai Wang, Feng Wang, Zhengyi Wang, and Huaping Liu.
\newblock V3d: Video diffusion models are effective 3d generators.
\newblock \emph{arXiv preprint arXiv:2403.06738}, 2024{\natexlab{a}}.

\bibitem[Chen et~al.(2024{\natexlab{b}})Chen, Yang, Huang, Lutio, Esturo, Ivanovic, Litany, Gojcic, Fidler, Pavone, Song, and Wang]{omnire}
Ziyu Chen, Jiawei Yang, Jiahui Huang, Riccardo~de Lutio, Janick~Martinez Esturo, Boris Ivanovic, Or Litany, Zan Gojcic, Sanja Fidler, Marco Pavone, Li Song, and Yue Wang.
\newblock Omnire: Omni urban scene reconstruction.
\newblock \emph{arXiv preprint arXiv:2408.16760}, 2024{\natexlab{b}}.

\bibitem[Cheng et~al.(2023)Cheng, Long, Yin, Wang, Wu, Ma, Wang, Chen, and Chen]{ucnerf}
Kai Cheng, Xiaoxiao Long, Wei Yin, Jin Wang, Zhiqiang Wu, Yuexin Ma, Kaixuan Wang, Xiaozhi Chen, and Xuejin Chen.
\newblock Uc-nerf: Neural radiance field for under-calibrated multi-view cameras in autonomous driving.
\newblock \emph{arXiv preprint arXiv:2311.16945}, 2023.

\bibitem[Fridovich-Keil et~al.(2023)Fridovich-Keil, Meanti, Warburg, Recht, and Kanazawa]{kplane}
Sara Fridovich-Keil, Giacomo Meanti, Frederik~Rahb{\ae}k Warburg, Benjamin Recht, and Angjoo Kanazawa.
\newblock K-planes: Explicit radiance fields in space, time, and appearance.
\newblock In \emph{CVPR}, 2023.

\bibitem[Gao et~al.(2024{\natexlab{a}})Gao, Chen, Li, Hong, Li, and Xu]{magicdrive3d}
Ruiyuan Gao, Kai Chen, Zhihao Li, Lanqing Hong, Zhenguo Li, and Qiang Xu.
\newblock Magicdrive3d: Controllable 3d generation for any-view rendering in street scenes.
\newblock \emph{arXiv preprint arXiv:2405.14475}, 2024{\natexlab{a}}.

\bibitem[Gao et~al.(2024{\natexlab{b}})Gao, Holynski, Henzler, Brussee, Martin-Brualla, Srinivasan, Barron, and Poole]{gao2024cat3d}
Ruiqi Gao, Aleksander Holynski, Philipp Henzler, Arthur Brussee, Ricardo Martin-Brualla, Pratul Srinivasan, Jonathan~T Barron, and Ben Poole.
\newblock Cat3d: Create anything in 3d with multi-view diffusion models.
\newblock \emph{arXiv preprint arXiv:2405.10314}, 2024{\natexlab{b}}.

\bibitem[Gao et~al.(2024{\natexlab{c}})Gao, Yang, Chen, Chitta, Qiu, Geiger, Zhang, and Li]{vista}
Shenyuan Gao, Jiazhi Yang, Li Chen, Kashyap Chitta, Yihang Qiu, Andreas Geiger, Jun Zhang, and Hongyang Li.
\newblock Vista: A generalizable driving world model with high fidelity and versatile controllability.
\newblock \emph{arXiv preprint arXiv:2405.17398}, 2024{\natexlab{c}}.

\bibitem[Girdhar et~al.(2023)Girdhar, Singh, Brown, Duval, Azadi, Rambhatla, Shah, Yin, Parikh, and Misra]{emuvideo}
Rohit Girdhar, Mannat Singh, Andrew Brown, Quentin Duval, Samaneh Azadi, Sai~Saketh Rambhatla, Akbar Shah, Xi Yin, Devi Parikh, and Ishan Misra.
\newblock Emu video: Factorizing text-to-video generation by explicit image conditioning.
\newblock \emph{arXiv preprint arXiv:2311.10709}, 2023.

\bibitem[Guo et~al.(2023)Guo, Deng, Li, Bai, Shi, Wang, Ding, Wang, and Li]{streetsurf}
Jianfei Guo, Nianchen Deng, Xinyang Li, Yeqi Bai, Botian Shi, Chiyu Wang, Chenjing Ding, Dongliang Wang, and Yikang Li.
\newblock Streetsurf: Extending multi-view implicit surface reconstruction to street views.
\newblock \emph{arXiv preprint arXiv:2306.04988}, 2023.

\bibitem[Gupta et~al.(2023)Gupta, Yu, Sohn, Gu, Hahn, Fei-Fei, Essa, Jiang, and Lezama]{gupta2023photorealistic}
Agrim Gupta, Lijun Yu, Kihyuk Sohn, Xiuye Gu, Meera Hahn, Li Fei-Fei, Irfan Essa, Lu Jiang, and Jos{\'e} Lezama.
\newblock Photorealistic video generation with diffusion models.
\newblock \emph{arXiv preprint arXiv:2312.06662}, 2023.

\bibitem[Han et~al.(2024)Han, Zhou, Long, Wang, and Xiao]{han2024ggs}
Huasong Han, Kaixuan Zhou, Xiaoxiao Long, Yusen Wang, and Chunxia Xiao.
\newblock Ggs: Generalizable gaussian splatting for lane switching in autonomous driving.
\newblock \emph{arXiv preprint arXiv:2409.02382}, 2024.

\bibitem[Heusel et~al.(2017)Heusel, Ramsauer, Unterthiner, Nessler, and Hochreiter]{fid}
Martin Heusel, Hubert Ramsauer, Thomas Unterthiner, Bernhard Nessler, and Sepp Hochreiter.
\newblock Gans trained by a two time-scale update rule converge to a local nash equilibrium.
\newblock \emph{NeurIPS}, 2017.

\bibitem[Ho et~al.(2022{\natexlab{a}})Ho, Chan, Saharia, Whang, Gao, Gritsenko, Kingma, Poole, Norouzi, Fleet, et~al.]{ho2022imagen}
Jonathan Ho, William Chan, Chitwan Saharia, Jay Whang, Ruiqi Gao, Alexey Gritsenko, Diederik~P Kingma, Ben Poole, Mohammad Norouzi, David~J Fleet, et~al.
\newblock Imagen video: High definition video generation with diffusion models.
\newblock \emph{arXiv preprint arXiv:2210.02303}, 2022{\natexlab{a}}.

\bibitem[Ho et~al.(2022{\natexlab{b}})Ho, Salimans, Gritsenko, Chan, Norouzi, and Fleet]{ho2022video}
Jonathan Ho, Tim Salimans, Alexey Gritsenko, William Chan, Mohammad Norouzi, and David~J Fleet.
\newblock Video diffusion models.
\newblock \emph{NeurIPS}, 2022{\natexlab{b}}.

\bibitem[Hong et~al.(2022)Hong, Ding, Zheng, Liu, and Tang]{hong2022cogvideo}
Wenyi Hong, Ming Ding, Wendi Zheng, Xinghan Liu, and Jie Tang.
\newblock Cogvideo: Large-scale pretraining for text-to-video generation via transformers.
\newblock \emph{arXiv preprint arXiv:2205.15868}, 2022.

\bibitem[Hu et~al.(2023{\natexlab{a}})Hu, Russell, Yeo, Murez, Fedoseev, Kendall, Shotton, and Corrado]{gaia}
Anthony Hu, Lloyd Russell, Hudson Yeo, Zak Murez, George Fedoseev, Alex Kendall, Jamie Shotton, and Gianluca Corrado.
\newblock Gaia-1: A generative world model for autonomous driving.
\newblock \emph{arXiv preprint arXiv:2309.17080}, 2023{\natexlab{a}}.

\bibitem[Hu et~al.(2022)Hu, Chen, Wu, Li, Yan, and Tao]{stp3}
Shengchao Hu, Li Chen, Penghao Wu, Hongyang Li, Junchi Yan, and Dacheng Tao.
\newblock St-p3: End-to-end vision-based autonomous driving via spatial-temporal feature learning.
\newblock In \emph{ECCV}, 2022.

\bibitem[Hu et~al.(2023{\natexlab{b}})Hu, Yang, Chen, Li, Sima, Zhu, Chai, Du, Lin, Wang, et~al.]{uniad}
Yihan Hu, Jiazhi Yang, Li Chen, Keyu Li, Chonghao Sima, Xizhou Zhu, Siqi Chai, Senyao Du, Tianwei Lin, Wenhai Wang, et~al.
\newblock Planning-oriented autonomous driving.
\newblock In \emph{CVPR}, 2023{\natexlab{b}}.

\bibitem[Huang et~al.(2024)Huang, Wei, Zheng, An, Lu, Zhan, Tomizuka, Keutzer, and Zhang]{s3gaussian}
Nan Huang, Xiaobao Wei, Wenzhao Zheng, Pengju An, Ming Lu, Wei Zhan, Masayoshi Tomizuka, Kurt Keutzer, and Shanghang Zhang.
\newblock $s^3$gaussian: Self-supervised street gaussians for autonomous driving.
\newblock \emph{arXiv preprint arXiv:2405.20323}, 2024.

\bibitem[Irshad et~al.(2023)Irshad, Zakharov, Liu, Guizilini, Kollar, Gaidon, Kira, and Ambrus]{neo360}
Muhammad~Zubair Irshad, Sergey Zakharov, Katherine Liu, Vitor Guizilini, Thomas Kollar, Adrien Gaidon, Zsolt Kira, and Rares Ambrus.
\newblock Neo 360: Neural fields for sparse view synthesis of outdoor scenes.
\newblock In \emph{ICCV}, 2023.

\bibitem[Jiang et~al.(2023)Jiang, Chen, Xu, Liao, Chen, Zhou, Zhang, Liu, Huang, and Wang]{vad}
Bo Jiang, Shaoyu Chen, Qing Xu, Bencheng Liao, Jiajie Chen, Helong Zhou, Qian Zhang, Wenyu Liu, Chang Huang, and Xinggang Wang.
\newblock Vad: Vectorized scene representation for efficient autonomous driving.
\newblock In \emph{ICCV}, 2023.

\bibitem[Jocher and Qiu(2024)]{yolo11_ultralytics}
Glenn Jocher and Jing Qiu.
\newblock Ultralytics yolo11, 2024.

\bibitem[Kerbl et~al.(2023)Kerbl, Kopanas, Leimk{\"u}hler, and Drettakis]{3dgs}
Bernhard Kerbl, Georgios Kopanas, Thomas Leimk{\"u}hler, and George Drettakis.
\newblock 3d gaussian splatting for real-time radiance field rendering.
\newblock \emph{ACM ToG}, 2023.

\bibitem[Kingma and Ba(2014)]{kingma2014adam}
Diederik~P Kingma and Jimmy Ba.
\newblock Adam: A method for stochastic optimization.
\newblock \emph{arXiv preprint arXiv:1412.6980}, 2014.

\bibitem[Kondratyuk et~al.(2023)Kondratyuk, Yu, Gu, Lezama, Huang, Hornung, Adam, Akbari, Alon, Birodkar, et~al.]{kondratyuk2023videopoet}
Dan Kondratyuk, Lijun Yu, Xiuye Gu, Jos{\'e} Lezama, Jonathan Huang, Rachel Hornung, Hartwig Adam, Hassan Akbari, Yair Alon, Vighnesh Birodkar, et~al.
\newblock Videopoet: A large language model for zero-shot video generation.
\newblock \emph{arXiv preprint arXiv:2312.14125}, 2023.

\bibitem[Kundu et~al.(2022)Kundu, Genova, Yin, Fathi, Pantofaru, Guibas, Tagliasacchi, Dellaert, and Funkhouser]{panopticnerf}
Abhijit Kundu, Kyle Genova, Xiaoqi Yin, Alireza Fathi, Caroline Pantofaru, Leonidas~J Guibas, Andrea Tagliasacchi, Frank Dellaert, and Thomas Funkhouser.
\newblock Panoptic neural fields: A semantic object-aware neural scene representation.
\newblock In \emph{CVPR}, 2022.

\bibitem[LeCun and Courant(2022)]{lecun2022jepa}
Yann LeCun and Courant.
\newblock A path towards autonomous machine intelligence version 0.9.2, 2022-06-27.
\newblock 2022.

\bibitem[Li et~al.(2024{\natexlab{a}})Li, Yuan, Zhang, Wu, Zhao, Song, Feng, Ding, Zhang, and Wang]{xld}
Hao Li, Ming Yuan, Yan Zhang, Chenming Wu, Chen Zhao, Chunyu Song, Haocheng Feng, Errui Ding, Dingwen Zhang, and Jingdong Wang.
\newblock Xld: A cross-lane dataset for benchmarking novel driving view synthesis.
\newblock \emph{arXiv preprint arXiv:2406.18360}, 2024{\natexlab{a}}.

\bibitem[Li et~al.(2021)Li, Niklaus, Snavely, and Wang]{li2021neural}
Zhengqi Li, Simon Niklaus, Noah Snavely, and Oliver Wang.
\newblock Neural scene flow fields for space-time view synthesis of dynamic scenes.
\newblock In \emph{CVPR}, 2021.

\bibitem[Li et~al.(2024{\natexlab{b}})Li, Yu, Lan, Li, Kautz, Lu, and Alvarez]{li2024ego}
Zhiqi Li, Zhiding Yu, Shiyi Lan, Jiahan Li, Jan Kautz, Tong Lu, and Jose~M Alvarez.
\newblock Is ego status all you need for open-loop end-to-end autonomous driving?
\newblock In \emph{CVPR}, 2024{\natexlab{b}}.

\bibitem[Li et~al.(2024{\natexlab{c}})Li, Zhang, Wu, Zhu, and Zhang]{hogaussian}
Zhuopeng Li, Yilin Zhang, Chenming Wu, Jianke Zhu, and Liangjun Zhang.
\newblock Ho-gaussian: Hybrid optimization of 3d gaussian splatting for urban scenes.
\newblock \emph{arXiv preprint arXiv:2403.20032}, 2024{\natexlab{c}}.

\bibitem[Lin et~al.(2023)Lin, Gao, Tang, Takikawa, Zeng, Huang, Kreis, Fidler, Liu, and Lin]{magic3d}
Chen-Hsuan Lin, Jun Gao, Luming Tang, Towaki Takikawa, Xiaohui Zeng, Xun Huang, Karsten Kreis, Sanja Fidler, Ming-Yu Liu, and Tsung-Yi Lin.
\newblock Magic3d: High-resolution text-to-3d content creation.
\newblock In \emph{CVPR}, 2023.

\bibitem[Lin et~al.(2022)Lin, Peng, Xu, Yan, Shuai, Bao, and Zhou]{lin2022efficient}
Haotong Lin, Sida Peng, Zhen Xu, Yunzhi Yan, Qing Shuai, Hujun Bao, and Xiaowei Zhou.
\newblock Efficient neural radiance fields for interactive free-viewpoint video.
\newblock In \emph{SIGGRAPH Asia}, 2022.

\bibitem[Lu et~al.(2023)Lu, Xu, Chen, Li, Lin, and Jiang]{lu2023urban}
Fan Lu, Yan Xu, Guang Chen, Hongsheng Li, Kwan-Yee Lin, and Changjun Jiang.
\newblock Urban radiance field representation with deformable neural mesh primitives.
\newblock In \emph{ICCV}, 2023.

\bibitem[Ma et~al.(2024)Ma, Wang, Jia, Chen, Liu, Li, Chen, and Qiao]{ma2024latte}
Xin Ma, Yaohui Wang, Gengyun Jia, Xinyuan Chen, Ziwei Liu, Yuan-Fang Li, Cunjian Chen, and Yu Qiao.
\newblock Latte: Latent diffusion transformer for video generation.
\newblock \emph{arXiv preprint arXiv:2401.03048}, 2024.

\bibitem[Mildenhall et~al.(2021)Mildenhall, Srinivasan, Tancik, Barron, Ramamoorthi, and Ng]{nerf}
Ben Mildenhall, Pratul~P Srinivasan, Matthew Tancik, Jonathan~T Barron, Ravi Ramamoorthi, and Ren Ng.
\newblock Nerf: Representing scenes as neural radiance fields for view synthesis.
\newblock \emph{CACM}, 2021.

\bibitem[M{\"u}ller et~al.(2022)M{\"u}ller, Evans, Schied, and Keller]{ngp}
Thomas M{\"u}ller, Alex Evans, Christoph Schied, and Alexander Keller.
\newblock Instant neural graphics primitives with a multiresolution hash encoding.
\newblock \emph{ACM ToG}, 2022.

\bibitem[Ost et~al.(2021)Ost, Mannan, Thuerey, Knodt, and Heide]{ost2021neural}
Julian Ost, Fahim Mannan, Nils Thuerey, Julian Knodt, and Felix Heide.
\newblock Neural scene graphs for dynamic scenes.
\newblock In \emph{CVPR}, 2021.

\bibitem[Park et~al.(2021)Park, Sinha, Hedman, Barron, Bouaziz, Goldman, Martin-Brualla, and Seitz]{hypernerf}
Keunhong Park, Utkarsh Sinha, Peter Hedman, Jonathan~T Barron, Sofien Bouaziz, Dan~B Goldman, Ricardo Martin-Brualla, and Steven~M Seitz.
\newblock Hypernerf: A higher-dimensional representation for topologically varying neural radiance fields.
\newblock \emph{arXiv preprint arXiv:2106.13228}, 2021.

\bibitem[Podell et~al.(2023)Podell, English, Lacey, Blattmann, Dockhorn, M{\"u}ller, Penna, and Rombach]{sdxl}
Dustin Podell, Zion English, Kyle Lacey, Andreas Blattmann, Tim Dockhorn, Jonas M{\"u}ller, Joe Penna, and Robin Rombach.
\newblock Sdxl: Improving latent diffusion models for high-resolution image synthesis.
\newblock \emph{arXiv preprint arXiv:2307.01952}, 2023.

\bibitem[Poole et~al.(2022)Poole, Jain, Barron, and Mildenhall]{dreamfusion}
Ben Poole, Ajay Jain, Jonathan~T Barron, and Ben Mildenhall.
\newblock Dreamfusion: Text-to-3d using 2d diffusion.
\newblock \emph{arXiv preprint arXiv:2209.14988}, 2022.

\bibitem[Ramesh et~al.(2022)Ramesh, Dhariwal, Nichol, Chu, and Chen]{ramesh2022hierarchical}
Aditya Ramesh, Prafulla Dhariwal, Alex Nichol, Casey Chu, and Mark Chen.
\newblock Hierarchical text-conditional image generation with clip latents.
\newblock \emph{arXiv preprint arXiv:2204.06125}, 2022.

\bibitem[Rematas et~al.(2022)Rematas, Liu, Srinivasan, Barron, Tagliasacchi, Funkhouser, and Ferrari]{urbannerf}
Konstantinos Rematas, Andrew Liu, Pratul~P Srinivasan, Jonathan~T Barron, Andrea Tagliasacchi, Thomas Funkhouser, and Vittorio Ferrari.
\newblock Urban radiance fields.
\newblock In \emph{CVPR}, 2022.

\bibitem[Rombach et~al.(2022)Rombach, Blattmann, Lorenz, Esser, and Ommer]{sd}
Robin Rombach, Andreas Blattmann, Dominik Lorenz, Patrick Esser, and Bj{\"o}rn Ommer.
\newblock High-resolution image synthesis with latent diffusion models.
\newblock In \emph{CVPR}, 2022.

\bibitem[Saharia et~al.(2022)Saharia, Chan, Saxena, Li, Whang, Denton, Ghasemipour, Gontijo~Lopes, Karagol~Ayan, Salimans, et~al.]{saharia2022photorealistic}
Chitwan Saharia, William Chan, Saurabh Saxena, Lala Li, Jay Whang, Emily~L Denton, Kamyar Ghasemipour, Raphael Gontijo~Lopes, Burcu Karagol~Ayan, Tim Salimans, et~al.
\newblock Photorealistic text-to-image diffusion models with deep language understanding.
\newblock \emph{NeurIPS}, 2022.

\bibitem[Sargent et~al.(2023)Sargent, Li, Shah, Herrmann, Yu, Zhang, Chan, Lagun, Fei-Fei, Sun, et~al.]{sargent2023zeronvs}
Kyle Sargent, Zizhang Li, Tanmay Shah, Charles Herrmann, Hong-Xing Yu, Yunzhi Zhang, Eric~Ryan Chan, Dmitry Lagun, Li Fei-Fei, Deqing Sun, et~al.
\newblock Zeronvs: Zero-shot 360-degree view synthesis from a single real image.
\newblock \emph{arXiv preprint arXiv:2310.17994}, 2023.

\bibitem[Song et~al.(2023)Song, Chen, Li, Chen, Chen, Yuan, Xu, and Geiger]{nerfplayer}
Liangchen Song, Anpei Chen, Zhong Li, Zhang Chen, Lele Chen, Junsong Yuan, Yi Xu, and Andreas Geiger.
\newblock Nerfplayer: A streamable dynamic scene representation with decomposed neural radiance fields.
\newblock \emph{IEEE TVCG}, 2023.

\bibitem[Sun et~al.(2020)Sun, Kretzschmar, Dotiwalla, Chouard, Patnaik, Tsui, Guo, Zhou, Chai, Caine, Vasudevan, Han, Ngiam, Zhao, Timofeev, Ettinger, Krivokon, Gao, Joshi, Zhang, Shlens, Chen, and Anguelov]{waymo}
Pei Sun, Henrik Kretzschmar, Xerxes Dotiwalla, Aurelien Chouard, Vijaysai Patnaik, Paul Tsui, James Guo, Yin Zhou, Yuning Chai, Benjamin Caine, Vijay Vasudevan, Wei Han, Jiquan Ngiam, Hang Zhao, Aleksei Timofeev, Scott Ettinger, Maxim Krivokon, Amy Gao, Aditya Joshi, Yu Zhang, Jonathon Shlens, Zhifeng Chen, and Dragomir Anguelov.
\newblock Scalability in perception for autonomous driving: Waymo open dataset.
\newblock In \emph{CVPR}, 2020.

\bibitem[Tancik et~al.(2022)Tancik, Casser, Yan, Pradhan, Mildenhall, Srinivasan, Barron, and Kretzschmar]{blocknerf}
Matthew Tancik, Vincent Casser, Xinchen Yan, Sabeek Pradhan, Ben Mildenhall, Pratul~P Srinivasan, Jonathan~T Barron, and Henrik Kretzschmar.
\newblock Block-nerf: Scalable large scene neural view synthesis.
\newblock In \emph{CVPR}, 2022.

\bibitem[Tonderski et~al.(2024)Tonderski, Lindstr{\"o}m, Hess, Ljungbergh, Svensson, and Petersson]{Neurad}
Adam Tonderski, Carl Lindstr{\"o}m, Georg Hess, William Ljungbergh, Lennart Svensson, and Christoffer Petersson.
\newblock Neurad: Neural rendering for autonomous driving.
\newblock In \emph{CVPR}, 2024.

\bibitem[Voleti et~al.(2024)Voleti, Yao, Boss, Letts, Pankratz, Tochilkin, Laforte, Rombach, and Jampani]{voleti2024sv3d}
Vikram Voleti, Chun-Han Yao, Mark Boss, Adam Letts, David Pankratz, Dmitry Tochilkin, Christian Laforte, Robin Rombach, and Varun Jampani.
\newblock Sv3d: Novel multi-view synthesis and 3d generation from a single image using latent video diffusion.
\newblock \emph{arXiv preprint arXiv:2403.12008}, 2024.

\bibitem[Wang et~al.(2023)Wang, Zhu, Huang, Chen, Zhu, and Lu]{drivedreamer}
Xiaofeng Wang, Zheng Zhu, Guan Huang, Xinze Chen, Jiagang Zhu, and Jiwen Lu.
\newblock Drivedreamer: Towards real-world-driven world models for autonomous driving.
\newblock \emph{arXiv preprint arXiv:2309.09777}, 2023.

\bibitem[Wang et~al.(2024{\natexlab{a}})Wang, Zhao, Liu, Wang, Zhao, Bao, Zhu, Zhang, and Wang]{egovid}
Xiaofeng Wang, Kang Zhao, Feng Liu, Jiayu Wang, Guosheng Zhao, Xiaoyi Bao, Zheng Zhu, Yingya Zhang, and Xingang Wang.
\newblock Egovid-5m: A large-scale video-action dataset for egocentric video generation.
\newblock \emph{arXiv preprint arXiv:2411.08380}, 2024{\natexlab{a}}.

\bibitem[Wang et~al.(2024{\natexlab{b}})Wang, Zhu, Huang, Wang, Chen, and Lu]{worlddreamer}
Xiaofeng Wang, Zheng Zhu, Guan Huang, Boyuan Wang, Xinze Chen, and Jiwen Lu.
\newblock Worlddreamer: Towards general world models for video generation via predicting masked tokens.
\newblock \emph{arXiv preprint arXiv:2401.09985}, 2024{\natexlab{b}}.

\bibitem[Wang et~al.(2024{\natexlab{c}})Wang, He, Fan, Li, Chen, and Zhang]{drivewm}
Yuqi Wang, Jiawei He, Lue Fan, Hongxin Li, Yuntao Chen, and Zhaoxiang Zhang.
\newblock Driving into the future: Multiview visual forecasting and planning with world model for autonomous driving.
\newblock In \emph{CVPR}, 2024{\natexlab{c}}.

\bibitem[Wu et~al.(2024)Wu, Mildenhall, Henzler, Park, Gao, Watson, Srinivasan, Verbin, Barron, Poole, et~al.]{reconfusion}
Rundi Wu, Ben Mildenhall, Philipp Henzler, Keunhong Park, Ruiqi Gao, Daniel Watson, Pratul~P Srinivasan, Dor Verbin, Jonathan~T Barron, Ben Poole, et~al.
\newblock Reconfusion: 3d reconstruction with diffusion priors.
\newblock In \emph{CVPR}, 2024.

\bibitem[Wu et~al.(2023)Wu, Liu, Luo, Zhong, Chen, Xiao, Hou, Lou, Chen, Yang, et~al.]{mars}
Zirui Wu, Tianyu Liu, Liyi Luo, Zhide Zhong, Jianteng Chen, Hongmin Xiao, Chao Hou, Haozhe Lou, Yuantao Chen, Runyi Yang, et~al.
\newblock Mars: An instance-aware, modular and realistic simulator for autonomous driving.
\newblock In \emph{ICAI}, 2023.

\bibitem[Xiang et~al.(2024)Xiang, Liu, Gu, Gao, Ning, Zha, Feng, Tao, Hao, Shi, et~al.]{xiang2024pandora}
Jiannan Xiang, Guangyi Liu, Yi Gu, Qiyue Gao, Yuting Ning, Yuheng Zha, Zeyu Feng, Tianhua Tao, Shibo Hao, Yemin Shi, et~al.
\newblock Pandora: Towards general world model with natural language actions and video states.
\newblock \emph{arXiv preprint arXiv:2406.09455}, 2024.

\bibitem[Xie et~al.(2023)Xie, Zhang, Li, Zhang, and Zhang]{snerf}
Ziyang Xie, Junge Zhang, Wenye Li, Feihu Zhang, and Li Zhang.
\newblock S-nerf: Neural radiance fields for street views.
\newblock \emph{arXiv preprint arXiv:2303.00749}, 2023.

\bibitem[Yan et~al.(2021)Yan, Zhang, Abbeel, and Srinivas]{yan2021videogpt}
Wilson Yan, Yunzhi Zhang, Pieter Abbeel, and Aravind Srinivas.
\newblock Videogpt: Video generation using vq-vae and transformers.
\newblock \emph{arXiv preprint arXiv:2104.10157}, 2021.

\bibitem[Yan et~al.(2024)Yan, Lin, Zhou, Wang, Sun, Zhan, Lang, Zhou, and Peng]{streetgaussian}
Yunzhi Yan, Haotong Lin, Chenxu Zhou, Weijie Wang, Haiyang Sun, Kun Zhan, Xianpeng Lang, Xiaowei Zhou, and Sida Peng.
\newblock Street gaussians for modeling dynamic urban scenes.
\newblock \emph{arXiv preprint arXiv:2401.01339}, 2024.

\bibitem[Yang et~al.(2023{\natexlab{a}})Yang, Ivanovic, Litany, Weng, Kim, Li, Che, Xu, Fidler, Pavone, et~al.]{emernerf}
Jiawei Yang, Boris Ivanovic, Or Litany, Xinshuo Weng, Seung~Wook Kim, Boyi Li, Tong Che, Danfei Xu, Sanja Fidler, Marco Pavone, et~al.
\newblock Emernerf: Emergent spatial-temporal scene decomposition via self-supervision.
\newblock \emph{arXiv preprint arXiv:2311.02077}, 2023{\natexlab{a}}.

\bibitem[Yang et~al.(2024{\natexlab{a}})Yang, Wen, Ma, Mei, Li, Wei, Lei, Fu, Cai, Dou, Shi, He, Liu, and Qiao]{yang2024drivearena}
Xuemeng Yang, Licheng Wen, Yukai Ma, Jianbiao Mei, Xin Li, Tiantian Wei, Wenjie Lei, Daocheng Fu, Pinlong Cai, Min Dou, Botian Shi, Liang He, Yong Liu, and Yu Qiao.
\newblock Drivearena: A closed-loop generative simulation platform for autonomous driving.
\newblock \emph{arXiv preprint arXiv:2408.00415}, 2024{\natexlab{a}}.

\bibitem[Yang et~al.(2023{\natexlab{b}})Yang, Chen, Wang, Manivasagam, Ma, Yang, and Urtasun]{unisim}
Ze Yang, Yun Chen, Jingkang Wang, Sivabalan Manivasagam, Wei-Chiu Ma, Anqi~Joyce Yang, and Raquel Urtasun.
\newblock Unisim: A neural closed-loop sensor simulator.
\newblock In \emph{CVPR}, 2023{\natexlab{b}}.

\bibitem[Yang et~al.(2024{\natexlab{b}})Yang, Gao, Zhou, Jiao, Zhang, and Jin]{deformablegs}
Ziyi Yang, Xinyu Gao, Wen Zhou, Shaohui Jiao, Yuqing Zhang, and Xiaogang Jin.
\newblock Deformable 3d gaussians for high-fidelity monocular dynamic scene reconstruction.
\newblock In \emph{CVPR}, 2024{\natexlab{b}}.

\bibitem[Yang et~al.(2024{\natexlab{c}})Yang, Teng, Zheng, Ding, Huang, Xu, Yang, Hong, Zhang, Feng, et~al.]{yang2024cogvideox}
Zhuoyi Yang, Jiayan Teng, Wendi Zheng, Ming Ding, Shiyu Huang, Jiazheng Xu, Yuanming Yang, Wenyi Hong, Xiaohan Zhang, Guanyu Feng, et~al.
\newblock Cogvideox: Text-to-video diffusion models with an expert transformer.
\newblock \emph{arXiv preprint arXiv:2408.06072}, 2024{\natexlab{c}}.

\bibitem[Yifan et~al.(2019)Yifan, Serena, Wu, {\"O}ztireli, and Sorkine-Hornung]{yifan2019differentiable}
Wang Yifan, Felice Serena, Shihao Wu, Cengiz {\"O}ztireli, and Olga Sorkine-Hornung.
\newblock Differentiable surface splatting for point-based geometry processing.
\newblock \emph{ACM TOG}, 2019.

\bibitem[Yu et~al.(2024{\natexlab{a}})Yu, Chen, Huang, Sattler, and Geiger]{mipgs}
Zehao Yu, Anpei Chen, Binbin Huang, Torsten Sattler, and Andreas Geiger.
\newblock Mip-splatting: Alias-free 3d gaussian splatting.
\newblock In \emph{CVPR}, 2024{\natexlab{a}}.

\bibitem[Yu et~al.(2024{\natexlab{b}})Yu, Wang, Yang, Wang, Xie, Cai, Cao, Ji, and Sun]{sgd}
Zhongrui Yu, Haoran Wang, Jinze Yang, Hanzhang Wang, Zeke Xie, Yunfeng Cai, Jiale Cao, Zhong Ji, and Mingming Sun.
\newblock Sgd: Street view synthesis with gaussian splatting and diffusion prior.
\newblock \emph{arXiv preprint arXiv:2403.20079}, 2024{\natexlab{b}}.

\bibitem[Zeng et~al.(2023)Zeng, Wei, Zheng, Zou, Wei, Zhang, and Li]{pixeldance}
Yan Zeng, Guoqiang Wei, Jiani Zheng, Jiaxin Zou, Yang Wei, Yuchen Zhang, and Hang Li.
\newblock Make pixels dance: High-dynamic video generation.
\newblock \emph{arXiv preprint arXiv:2311.10982}, 2023.

\bibitem[Zhai et~al.(2023)Zhai, Feng, Du, Mao, Liu, Tan, Zhang, Ye, and Wang]{admlp}
Jiang-Tian Zhai, Ze Feng, Jinhao Du, Yongqiang Mao, Jiang-Jiang Liu, Zichang Tan, Yifu Zhang, Xiaoqing Ye, and Jingdong Wang.
\newblock Rethinking the open-loop evaluation of end-to-end autonomous driving in nuscenes.
\newblock \emph{arXiv preprint arXiv:2305.10430}, 2023.

\bibitem[Zhao et~al.(2024)Zhao, Wang, Zhu, Chen, Huang, Bao, and Wang]{drivedreamer2}
Guosheng Zhao, Xiaofeng Wang, Zheng Zhu, Xinze Chen, Guan Huang, Xiaoyi Bao, and Xingang Wang.
\newblock Drivedreamer-2: Llm-enhanced world models for diverse driving video generation.
\newblock \emph{arXiv preprint arXiv:2403.06845}, 2024.

\bibitem[Zhou et~al.(2024)Zhou, Lin, Shan, Wang, Sun, and Yang]{drivinggaussian}
Xiaoyu Zhou, Zhiwei Lin, Xiaojun Shan, Yongtao Wang, Deqing Sun, and Ming-Hsuan Yang.
\newblock Drivinggaussian: Composite gaussian splatting for surrounding dynamic autonomous driving scenes.
\newblock In \emph{CVPR}, pages 21634--21643, 2024.

\bibitem[Zhu et~al.(2024)Zhu, Wang, Zhao, Min, Deng, Dou, Wang, Shi, Wang, Zhang, et~al.]{zhu2024sora}
Zheng Zhu, Xiaofeng Wang, Wangbo Zhao, Chen Min, Nianchen Deng, Min Dou, Yuqi Wang, Botian Shi, Kai Wang, Chi Zhang, et~al.
\newblock Is sora a world simulator? a comprehensive survey on general world models and beyond.
\newblock \emph{arXiv preprint arXiv:2405.03520}, 2024.

\end{thebibliography}
}

\clearpage

\twocolumn[{%
\begin{center}
\centering
\includegraphics[width=\textwidth]{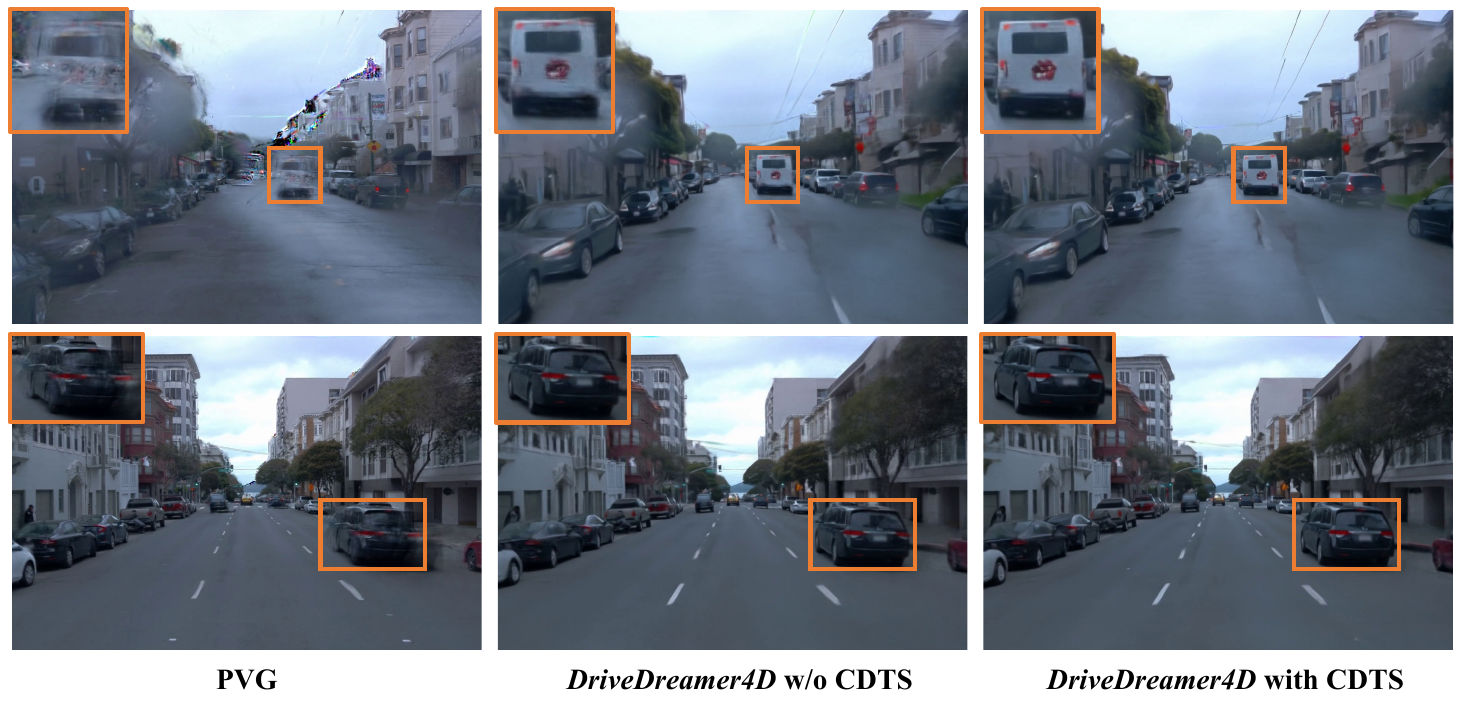}
\captionof{figure}{Visual comparisons in the novel trajectories for the Cousin Data Training Strategy (CDTS) ablation study. The \textcolor{orange}{orange} boxes emphasize the superior performance of \textit{DriveDreamer4D} and the further improvements in detail rendering brought by CDTS.}
\label{fig:ablation}
\end{center}}]

\begin{figure*}[!ht] %变速新视角
\centering
\includegraphics[width=\textwidth]{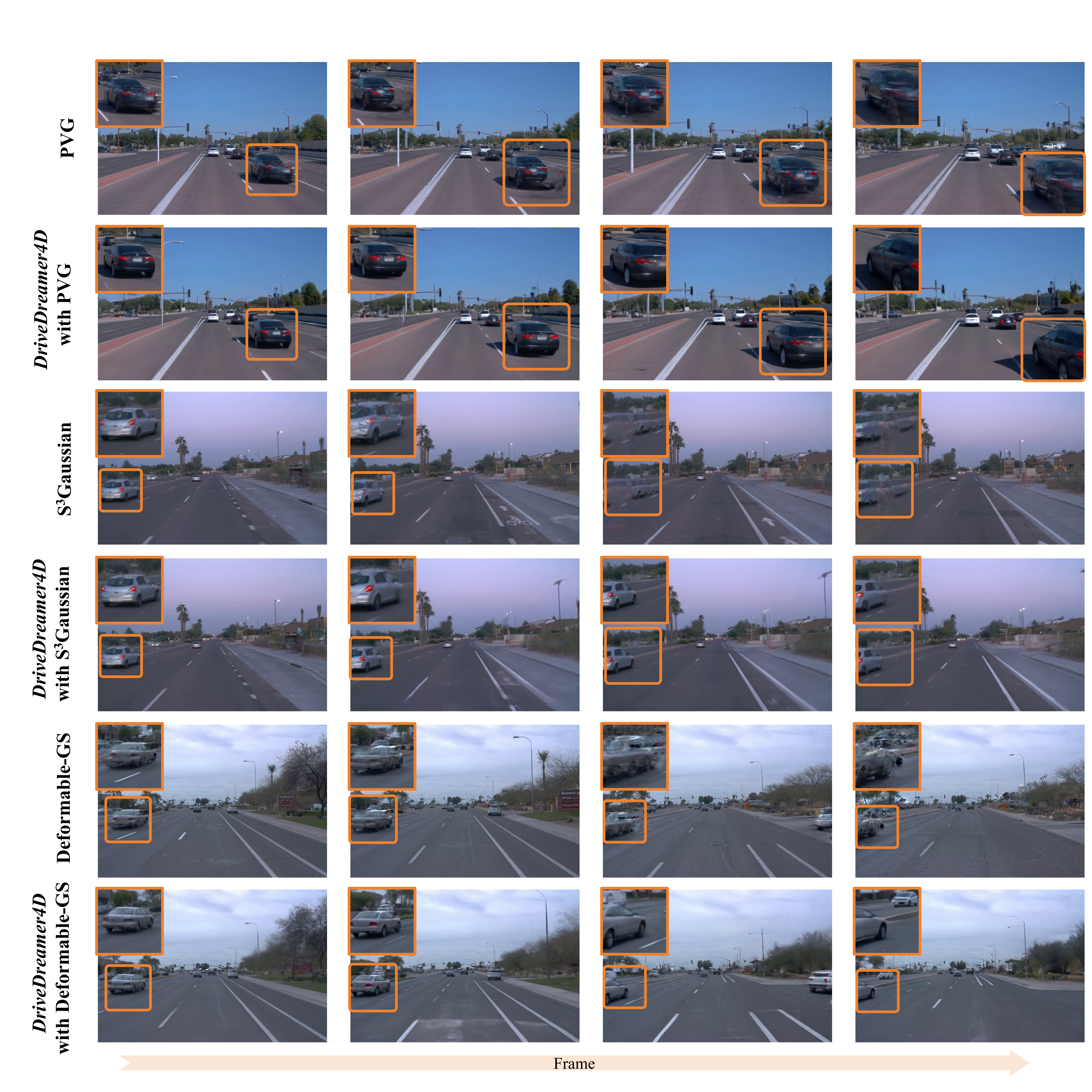}
\caption{Qualitative comparisons of novel trajectory renderings during speed change scenarios. The \textcolor{orange}{orange} boxes highlight that \textit{DriveDreamer4D} significantly enhances the rendering quality across various baseline methods (PVG \cite{pvg}, $\text{S}^3$Gaussian \cite{s3gaussian}, Deformable-GS \cite{deformablegs}).}
\label{fig:speed}
%\vspace{-1.5em}
\end{figure*}

In the supplementary material, we begin by introducing the three baseline methods employed in our work. Next, we elaborate on the implementation of \textit{DriveDreamer4D}, covering the training for novel trajectory video generation, the selection of scenes, and the setup of the user study. Finally, additional visualizations are presented to illustrate the improved rendering quality achieved through Cousin Data Training Strategy (CDTS) and showcase the performance of \textit{DriveDreamer4D} in speed change scenarios.

\begin{table*}
    \centering
    \begin{tabular}{lcc}
        \toprule
        \makecell[c]{Scene} & Start Frame & End Frame \\ \midrule
        segment-10359308928573410754\_720\_000\_740\_000\_with\_camera\_labels.tfrecord & 120&159\\
        segment-12820461091157089924\_5202\_916\_5222\_916\_with\_camera\_labels.tfrecord & 0 &39\\
        segment-15021599536622641101\_556\_150\_576\_150\_with\_camera\_labels.tfrecord&0&39\\
        segment-16767575238225610271\_5185\_000\_5205\_000\_with\_camera\_labels.tfrecord&0&39\\
        segment-17152649515605309595\_3440\_000\_3460\_000\_with\_camera\_labels.tfrecord&60&99\\
        segment-17860546506509760757\_6040\_000\_6060\_000\_with\_camera\_labels.tfrecord&90&129\\
        segment-2506799708748258165\_6455\_000\_6475\_000\_with\_camera\_labels.tfrecord&80&119\\
        segment-3015436519694987712\_1300\_000\_1320\_000\_with\_camera\_labels.tfrecord&40&79\\
        \bottomrule
        \end{tabular}
        \caption{Selected scenes from the validation set of the Waymo dataset \cite{waymo}.}
        \label{tab:scene}
\end{table*}

\section{Baselines}

To demonstrate the effectiveness and generalizability of our method, three different 4D Gaussian Splatting (4DGS) baselines are selected for the experiments. In this section, we briefly introduce the three baselines employed in this paper: PVG \cite{pvg}, $\text{S}^3$Gaussian \cite{s3gaussian}, and Deformable-GS \cite{deformablegs}.

\noindent \textbf{PVG} \cite{pvg} introduces a unified representation model known as Periodic Vibration Gaussians (PVGs), which vibrate over time with optimizable parameters, including vibration directions, lifespan, and life peak (the moment of highest opacity), to effectively represent dynamic scenes. The model employs a self-supervised approach to optimize these Gaussians and achieves static-dynamic decomposition by classifying them based on their lifespans. This method allows PVG to effectively represent the characteristics of various objects and elements in dynamic urban scenes.

\noindent \textbf{$\text{S}^3$Gaussian} \cite{s3gaussian} proposes a self-supervised street Gaussian method to model complex 4D dynamic scenes. Each scene is represented using 3D Gaussians to preserve explicitness, and a spatial-temporal field network is employed to compactly model the 4D dynamics. To facilitate efficient scene reconstruction without costly annotations, it utilizes an self-supervised approach to decompose dynamic and static 3D Gaussians.

\noindent \textbf{Deformable-GS} \cite{deformablegs} represents scenes using a canonical space defined by Gaussian distributions. It models scene dynamic by employing a deformation network to predict offsets for the Gaussian parameters. These offsets adjust the Gaussians to align with the dynamic elements of the scene. Additionally, Deformable-GS has demonstrated strong performance in both synthetic and indoor datasets.

\section{Implementation Details} 
\label{sec_sup_2}

In Sec. \ref{sec_sup_2}, we primarily introduce the training for novel trajectory video generation, the selection of scenes, and the details of the user study.

\noindent \textbf{Training for Novel Trajectory Video Generation.} As depicted in the upper part of Fig.~\ref{fig_framework} (in the main text), a controllable driving video generation model is crucial for producing novel trajectory videos. Specifically, we follow the approach outlined in \cite{drivedreamer2} to train such a model on the Waymo dataset \cite{waymo}. Unlike \cite{drivedreamer2}, which focuses on multi-view video generation using the nuScenes dataset \cite{nusc}, our work concentrates solely on front-view video generation. This focus allows us to increase the number of frames to 40 and the resolution to $960 \times 640$, a significant improvement compared to the previous 8 frames at a resolution of $448 \times 256$. The increase in both frame count and resolution contributes to an enhanced performance of the reconstruction model, particularly for novel trajectory generation. As for the training data, it comprises the entire Waymo training split, consisting of 798 videos. To enhance the dataset, we further divide these videos into 40-frame clips, resulting in approximately 64K clips. Additionally, the training process is initialized with parameters from SVD \cite{svd}, with 3D bounding boxes, HDMaps, and text incorporated as control conditions. And, the AdamW optimizer \cite{kingma2014adam} is employed for parameter optimization, with a learning rate of $5 \times 10^{-5}$, a batch size of 8, and a total of 50K iterations. All experiments are conducted on an NVIDIA H20 (96GB) GPUs.

\noindent \textbf{Scene Selection.} All selected scenes are sourced from the validation set of the Waymo dataset \cite{waymo} and are carefully chosen based on their distinctive characteristics. Specifically, the selection prioritizes scenes that exhibit significant motion dynamics, such as large-scale maneuvers, as these scenarios pose greater challenges for both video reconstruction and trajectory generation tasks. Tab.~\ref{tab:scene} shows all 8 scenes selected for our experiments. The official file names of these scenes, as provided in \cite{waymo}, are listed along with their respective starting and ending frames.

\noindent \textbf{User Study.} For the eight different scenes mentioned above, we create 72 comparison videos for the user study, covering three novel trajectories (acceleration, deceleration, and lane change) under three different baselines. To ensure fairness, the baseline and our method were randomly assigned to the left or right side of each comparison video. For each comparison, the participants are asked to choose the result they deem the most accurate or realistic (either the left or right side).

\section{Visualization}

In this part, we present additional visualization results, including qualitative analyses from the Cousin Data Training Strategy (CDTS) ablation study and visual comparisons for speed change scenarios.

As mentioned in Sec.~\ref{sec_4.3} of the main text, we perform an ablation study on the CDTS using PVG \cite{pvg}. For clarity, \textit{DriveDreamer4D} in this ablation study refers to \textit{DriveDreamer4D} with PVG. As shown in Fig.~\ref{fig:ablation}, \textit{DriveDreamer4D} demonstrates significant improvement over the baseline methods, regardless of whether CDTS is applied. Notably, the baseline methods struggle to accurately reconstruct the positions of vehicles in novel trajectories, resulting in severe ghosting artifacts. In contrast, \textit{DriveDreamer4D} excels at rendering the vehicle positions with high precision, significantly enhancing rendering performance. Moreover, with the introduction of CDTS, \textit{DriveDreamer4D} further enhances the reconstruction quality of dynamic vehicles, particularly at the edges, providing more detailed and accurate representations.

In Sec.~\ref{sec4.2} of the main text, we analyze the improved visualization effects of \textit{DriveDreamer4D} in lane change scenarios. For more details, please refer to the file videos/lane\_change\_comparison.mp4. More qualitative analysis of novel trajectory view renderings are shown in Fig.~\ref{fig:speed}, focusing on speed change scenarios. Our method significantly enhances the positional accuracy of foreground vehicles and background elements under speed change scenarios. Specifically, baseline results (PVG \cite{pvg}, $\text{S}^3$Gaussian \cite{s3gaussian}, Deformable-GS \cite{deformablegs}) are displayed in rows 1, 3, and 5. It is evident that the baseline methods face challenges with perspective synthesis in speed-change scenarios, resulting in inaccurate positional shifts (such as blurring or disappearance of foreground vehicles). In contrast, the integration of \textit{DriveDreamer4D} enables the 4DGS algorithms to achieve superior spatial consistency and significantly improved rendering quality, as illustrated by the orange boxes in the Fig.~\ref{fig:speed}. More details can be found in the file videos/speed\_change\_comparison.mp4.

% WARNING: do not forget to delete the supplementary pages from your submission 
% \input{sec/X_suppl}

\end{document}